\documentclass[accepted]{uai2022} 

\usepackage[american]{babel}

\usepackage{natbib} 
\usepackage{mathtools} 
\usepackage{booktabs} 
\usepackage{tikz} 

\usepackage{times}
\usepackage{helvet}
\usepackage{courier}
\usepackage{mathrsfs}
\usepackage{amssymb}
\usepackage{bbm}
\usepackage{dsfont}
\usepackage{bbold}
\usepackage{mathbbol}
\usepackage{newtxmath}
\usepackage{xcolor}

\usepackage{graphicx}
\usepackage{subfigure}
\usepackage{amsmath}
\usepackage{multirow}
\usepackage{wrapfig}
\usepackage{appendix}
\usepackage{color}

\usepackage{url}            
\usepackage{booktabs}       
\usepackage{amsfonts}       
\usepackage{nicefrac}       
\usepackage{microtype}      
\usepackage{graphicx}
\usepackage{subfigure}

\usepackage{algorithm}
\usepackage{algorithmic}

\usepackage{newfloat}
\usepackage{listings}
\def\firstauthor#1{\gdef\@firstauthor{#1}}

\title{MEPG: A Minimalist Ensemble Policy Gradient Framework for Deep Reinforcement Learning}

\author[1,2]{Qiang He}
\author[3]{Huangyuan Su}
\author[1,2]{Chen Gong}
\author[1]{Xinwen Hou}

\affil[1]{%
    Institute of Automation\\
    Chinese Academy of Sciences\\
    Beijing, China
}
\affil[2]{%
    School of Artificial Intelligence\\
    University of Chinese Academy of Sciences\\
    Beijing, China
}
\affil[3]{%
    School of Computer Science\\
    Carnegie Mellon University\\
    Pittsburgh, United States
}

\begin{document}
\maketitle

\begin{abstract}
During the training of a reinforcement learning (RL) agent, the distribution of training data is non-stationary as the agent's behavior changes over time. Therefore, there is a risk that the agent is overspecialized to a particular distribution and its performance suffers in the larger picture. Ensemble RL can mitigate this issue by learning a robust policy. However, it suffers from heavy computational resource consumption due to the newly introduced value and policy functions. In this paper, to avoid the notorious resources consumption issue, we design a novel and simple ensemble deep RL framework that integrates multiple models into a single model. Specifically, we propose the \underline{M}inimalist \underline{E}nsemble \underline{P}olicy \underline{G}radient framework (MEPG), which introduces minimalist ensemble consistent Bellman update by utilizing a modified dropout operator. MEPG holds ensemble property by keeping the dropout consistency of both sides of the Bellman equation. Additionally, the dropout operator also increases MEPG's generalization capability. Moreover, we theoretically show that the policy evaluation phase in the MEPG maintains two synchronized deep Gaussian Processes. To verify the MEPG framework's ability to generalize, we perform experiments on the gym simulator, which presents that the MEPG framework outperforms or achieves a similar level of performance as the current state-of-the-art ensemble methods and model-free methods without increasing additional computational resource costs.
\footnotetext{\textit{Decision Awareness in Reinforcement Learning Workshop at the 39th International Conference on Machine Learning (ICML)}, Baltimore, Maryland, USA, 2022. Copyright 2022 by the author(s).}

\end{abstract}

\section{Introduction}\label{sec: introduction}
One of the main driving factors for the success of the deep learning research is that intelligent systems can obtain open-world perceptions from large amounts of data \citep{brown2020language, he2016deep} through high-capacity function approximators.  Deep reinforcement learning (DRL) follows this paradigm, using neural networks for function approximation to solve sequential decision-making problems, and has achieved great success in a wide range of domains such as board games \citep{alpha}, video games \citep{nature_dqn, starcraft}, robot manipulation \citep{sac}, etc.

 DRL algorithms are often criticised for their drastic decrease of performance when tested in unfamiliar environments or domains\citep{kirk2021survey}. Even in the same environment, the agent can fail to adapt to different initiations of the environment. This is because the sequence of value approximation problems faced by the RL agent is not stationary\citep{dabney2020value}. As its policy improves, the distribution of the states and their values is also changing. Thus, a single agent can be overfitted to certain distributions. Ensemble methods can mitigate this problem by aggregating all the encountered scenarios collected from multiple models\citep{mesbah2021domain,wiering2008ensemble,osband2016deep, chua2018deep,kurutach2018model, saphal2021seerl, sunrise}. However, they introduce additional networks and loss functions which translate to a huge computational burden. Therefore, it is a challenge to create a DRL agent which maintains both generalizability and low computation consumption \citep{drl_matters, andrychowicz2021matters}. Ensemble RL methods, however, bring about heavy computational resource consumption issues due to multiple networks used. The ensemble methods in DRL work well when the models are sufficiently rich in diversity. Furthermore, additional networks also introduce more hyper-parameters, and requires more non-algorithmic level tricks and subtle fine-tuning for hyper-parameters. Due to these requirements, there are still many issues to be addressed when applying ensemble methods to DRL algorithms in practice.

Therefore, we ask the following question: can we find a simple approach to ensemble DRL algorithms to tackle the heavy computational resource consumption issue? Our answer is that one network is enough. In this paper, we propose the \underline{M}inimalist \underline{E}nsemble \underline{P}olicy \underline{G}radient framework (MEPG) to deal with the aforementioned issues of ensemble methods in DRL. Our insight origins from the fact that ensemble methods can be achieved by integrating multiple models into a single model. In this way, the issue of heavy resource consumption of ensemble learning methods can be well addressed, as one single model does not consume more computational resources than model-free deep RL algorithms. In addition, this framework introduces only one additional hyper-parameter on modern model-free DRL algorithms. We implement our insight by minimalist ensemble consistent Bellman update that utilizes modified dropout operator\citep{srivastava2014dropout}. The ensemble property holds as follows. The complete neural network being acted on by sampled dropout operator is equivalent to a new sub-model (or sub-network) in the training phase. In the inference phase, the complete model is used. And the complete model without the dropout operator is equivalent to all the sub-networks acting together. However, applying the dropout operator directly to the value functions creates the problem of source-target mismatch, i.e., the left and right sides of the Bellman equation do not correspond to the same value function. And a bad value function is learned, thus leading to limited policy improvement (check section \ref{sec: ablation study} for more details). Therefore, we introduce a minimalist ensemble consistent Bellman update where the same dropout operator acts on both sides of the Bellman equation at the same time. Furthermore, we show theoretically that the policy evaluation process in the MEPG framework maintains two synchronized deep Gaussian Processes \citep{damianou2013deep}. The consistency introduced by MEPG naturally improves the stability of Bellman equation compared to random mask situations. We apply MEPG to DRL algorithms DDPG \citep{ddpg} and SAC \citep{sac}. Our experimental results show that our modified algorithms outperform state-of-the-art ensemble DRL and model-free DRL algorithms. We highlight that our framework does not introduce any auxiliary loss function or additional computational consumption compared to the conventional model-free algorithms DDPG\citep{ddpg} and SAC\citep{sac}. Moreover, the parameters of MEPG are much less than those of the modern model-free DRL algorithms.

Our contributions are summarized as follows. First, we propose a general ensemble RL framework, called MEPG, which is simple and easy to implement. Unlike conventional ensemble Deep RL methods, this framework does not need to introduce any additional loss functions and computational costs. This framework can be combined with any DRL algorithm. Second, we provide theoretical analysis that shows the policy evaluation process in the MEPG framework maintains two synchronized deep GPs. Third, we experimentally demonstrate the effectiveness of the MEPG framework by combining our algorithm with the DDPG and SAC algorithms. The results show that MEPG achieves or exceeds state-of-the-art ensemble method DRL with 14\% to 27\% of parameters and 10\% to 20\% of training time compared to ensemble RL methods ACE \citep{zhang2019ace}, SUNRISE \citep{sunrise}, and REDQ\citep{redq}. 

\begin{figure*}[t]
	
	\centering
	\includegraphics[width=\textwidth]{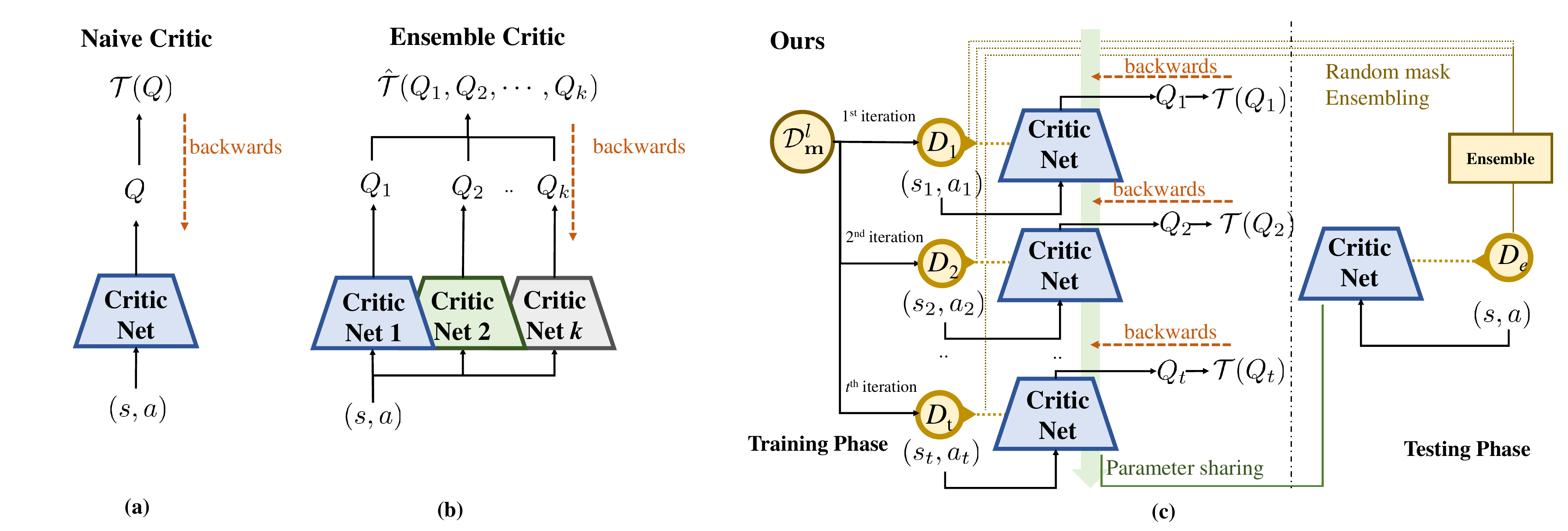}
	\caption{Neural network architectures for different critic. (a) is vanilla critic architectures. (b) is conventional ensemble critic architectures, which truly utilize multiple heterogeneous action value functions. (c) is our MEPG critic architecture. In the left part of (c), we train critic with minimalist ensemble consistent Bellman. In the right part of (c), the critic utilize a complete ensemble model.}
	\label{fig: structure}
\end{figure*}

\section{Related Work}
We briefly discuss off-policy DRL algorithms and ensemble DRL methods, and the dropout operator in DRL algorithms. Then we compare previous works with MEPG.
\subsection{Off-policy DRL algorithms} 
 Deep Q-networks (DQN) \citep{dqn} is the first successful applied off-policy deep RL algorithm. It has long been recognized that the overestimation in Q-learning could severely impair the performance \citep{thrun1993issues}. Double Q-learning \citep{hasselt2010double,ddqn} is proposed to solve overestimation issue for discrete action space. The overestimation issue still exists in continuous control algorithms such as Deterministic Policy Gradient (DPG) \citep{dpg} and its variant Deep Deterministic Policy Gradient (DDPG) \citep{ddpg}. \citep{td3} introduced clipped double Q-learning (CDQ), which decreases the overestimation issue in DDPG. \citep{sac} proposed Soft Actor Critic (SAC) algorithm which is based on maximum entropy RL framework \citep{ziebart2010modeling} and combined with CDQ, resulting in a stronger algorithm. Maximum entropy RL framework encourages exploration capacity by adding policy entropy to optimization objectives. The CDQ approach introduces a slight underestimation issue \citep{lan2019maxmin,he2020popo, he2020wd3}. We apply the MEPG framework to DDPG and SAC algorithm, which achieves or surpasses the performance of compared methods even without introducing a technique to get a precise value function.

\subsection{Ensemble DRL algorithms}
Ensemble methods use multiple learning algorithms to obtain better performance. They are also used in DRL \citep{zhang2019ace,wiering2008ensemble,osband2016deep, chua2018deep} for different purposes. \citep{kurutach2018model} showed that modeling error can be reduced by ensemble methods in model-based DRL. \citep{lan2019maxmin, redq} tackled the estimation issue and gave unbiased estimation methods of the value function from the perspective of ensemble functions. \citep{rem} proposed Random Ensemble Mixture, which introduces a convex combination of multiple Q-networks to approximate the optimal Q-function. \citep{saphal2021seerl} proposed a method for model training and selection in a single run. \citep{sunrise} proposed SUNRISE framework that uses an ensemble-based weighted Bellman backups and UCB exploration \citep{auer2002finite}. Unlike these methods, we only use a single model instead of multiple models to achieve our ensemble purpose.
\subsection{Dropout in DRL algorithms}

It was recognized earlier that dropout \citep{srivastava2014dropout} can improve the performance of DRL algorithms in video games \citep{lample2017playing,vinyals2019grandmaster}. Because the dropout operator prevents complex co-adaptations of units in neural networks where a feature detector is only helpful in the context of several other feature detectors. Therefore, the effectiveness of the dropout in pixel-level input DRL is similar to the supervised learning tasks for image input. Our work extends the dropout method in high dimensional inputs settings to low dimension input scenarios, where the effect of preventing pixel-level features co-adaptations may not exist. \citep{kamienny2020privileged} proposed a privileged information dropout method to improve sample efficiency and performance, but requires prior knowledge, i.e., an optimal representation for inputs. Our work does not require any auxiliary information. \citep{ICRA2019SafeRL,wu2021uncertainty} obtain the neural networks model uncertainty with dropout through multiple forwards given the same inputs. The closest work to ours is DQN+dropout \citep{gal2016dropout} in a discrete environment, which also utilizes the uncertainty of Q-function induced by dropout. MEPG is different from \citep{gal2016dropout} for the following reasons. First, MEPG maintains consistency of the two sides of Bellman equation where \citep{gal2016dropout} does not. Failure to maintain consistency leads to a mismatch in the Bellman equation and then harms performance (see Table \ref{table: ablation for me-ddpg 4 env}). Second, we use dropout for ensemble to improve performance and efficiency. \citep{gal2016dropout} uses dropout to obtain uncertainty thus using Thompson sampling to speed up convergence, without performance gain. Another approach to measuring uncertainty is to determine it using the variance generated by multiple Q-functions \citep{sunrise}. Our approach differs from previous works in that we neither use multiple Q-functions to estimate model uncertainty nor prevent pixel-level feature co-adaptation. We use the model ensemble property introduced by the dropout operator.

%

\section{Background}
We consider standard RL paradigm as a Markov Decision Process (MDP), which is characterized by a 6-tuple $(\mathcal{S, A, R, } P,\rho_0 ,\gamma)$, i.e., a state space $\mathcal{S}$, an action space $\mathcal{A}$, a reward function $r: \mathcal{S} \times \mathcal{A} \to \mathcal{R}$, a transition probability $P(s_{t+1} \mid s_t, a_t)$ - specifying the probability of transitioning from state $s_t$ to $s_{t+1}$ given action $a_t$, an initial state distribution $\rho_0$, and a discount factor $\gamma \in [0,1)$ \citep{rl}. The agent learns a policy, stochastic or deterministic by interacting with environment. At each time step, the agent generates action $a$ w.r.t. policy $\pi$ based on current state $s$ and send $a$ to environment. Then the agent receives a reward signal $r$ and a new state $s'$ from environment. Through multiple interactions, a trajectory $\tau = \{s_0, a_0, r_0, s_1, a_1, r_1, \cdots\}$ is generated. The optimization goal of RL is to maximize the expected cumulative discounted reward $J = \mathbb{E}_{\tau} [R_0]$, where $R_t = \sum_{i=t}^T \gamma^{i-t} r(s_i,a_i)$.   Two functions play important roles in RL, the state value function $V^\pi(s) = \mathbb{E}_\tau  [ R_0 | s_0=s ]$ and action value function (Q-function), $Q^\pi (s, a) = \mathbb{E}_\tau [ R_0 |s_0=s,a_0=a]$, a.k.a. critic, which is able to judge how good a state is and how good a specific action is respectively. According to Bellman Equation \citep{bellman1954theory}, action value function satisfies
\begin{equation}
Q^{\pi}(s,a) = r(s,a) + \gamma \mathbb{E}_{s' \sim P(\cdot|s,a), a'\sim \pi(\cdot \mid s')}[Q^\pi(s',a')].
\label{eq: bellman equation Q}
\end{equation}
The value function can be learned with Equation (\ref{eq: bellman equation Q}) \citep{rl}. For a large state space $\mathcal{S}$, we can utilize function approximation tools (e.g. neural networks) to represent the corresponding policy and Q-function. It is necessary to introduce the Policy Gradient Theorem \citep{policy_gradient} to learn a policy w.r.t. its return or value functions. In an actor-critic style algorithm, the critic is usually used to evaluate the quality of learning policy $\pi$. In RL literature, how to learn a critic is called policy evaluation and how to make a policy better is called policy improvement.

\section{Minimalist Ensemble Policy Gradient}
We first introduce minimalist ensemble consistent Bellman update and propose the minimalist ensemble policy gradient (MEPG) framework. Second, we apply MEPG to model-free off-policy DRL algorithms DDPG and SAC, resulting in ME-DDPG and ME-SAC. In principle, our framework can be applied to any modern DRL algorithm. Third, we show that the policy evaluation process in the MEPG framework theoretically maintains two synchronized deep GPs.
\subsection{Minimalist ensemble consistent Bellman update}
Computational resource consumption is a great challenge for the application of ensemble RL. Dropout operator has ensemble nature\citep{hara2016analysis} and reduces computational resource consumption. Therefore, we consider the dropout operator to be applied to ensemble RL. Specifically, we consider integrating $2^n$ sub-models into a single model. We deploy the dropout operator \citep{srivastava2014dropout} in our framework. The nature of the integration is guaranteed as follows. In the training phase, the neural network is equated to a sub-network after being acted upon by the sampled dropout operator. In the inference phase, the complete network that is not acted upon by the dropout operator is used, and the complete network is then equivalent to an ensemble network. Specifically, in the MEPG framework, the value network is trained under the above ensemble mechanism. When training the policy network, a complete value network (i.e., the ensemble value network) is used to induce the policy improvement phase. A feed-forward operation of a standard neural network for layer $l$ and hidden unit $i$ can be described as 
\begin{equation}
\begin{aligned}
z_i^{l+1} = \mathbf{w}_i^{l+1} \mathbf{x}^l + b_i^{l+1}, \quad 
x_i^{l+1} = f(z_i^{l+1}),
\end{aligned}
\end{equation}
where $f$ is an activation function, and $\mathbf{w}$ and $b$ represent the weights and biases respectively. We adopt the following dropout feed-forward style operation 
\begin{equation}
\begin{aligned}
&m_j^l\sim \text{Bernoulli}(1-p), \quad \tilde{\mathbf{x}}^l = \mathbf{m}^l \odot \mathbf{x}^l, \\
&z_i^{l+1} = f\big( \frac{1}{1-p} (\mathbf{w}_i^{l+1} \tilde{\mathbf{x}}^l + b_i^{l+1})\big),
\end{aligned}
\end{equation}
where $p$ is the probability of an element to be set to zero and $\odot$ represents Hadamard product. The scale factor $\frac{1}{1-p}$ is added to ensure that the expected output from each unit would keep the same as the one without the dropout operator $\mathcal{D}_\mathbf{m}^{l}$. However, if the dropout operator directly acts on Bellman equation in this form, the algorithm cannot learn a precise Q-function due to the mismatch between the left and the right sides of Bellman equation (\ref{eq: bellman equation Q}). As a result, the algorithm fails to learn the value function, i.e., fail to estimate the policy $\pi$, failing the whole process. To tackle this issue, we introduce minimalist ensemble consistent Bellman update. Let $\mathcal{D}_\mathbf{m}^l$ be a dropout operator acting on layer $l$ with parameter $\mathbf{m} \sim \text{Bernoulli}(1-p)$. We define the form of minimalist ensemble consistent Bellman update as
\begin{equation}
\begin{aligned}
\mathcal{D}_\mathbf{m}^{l} J^Q (\theta) = & \mathbb{E}_{(s,a)\sim \mathcal{B}} \Big[ \frac{1}{2} \Big(\mathcal{D}_\mathbf{m}^lQ(s,a; \theta) - \\
&\big(r(s,a) + \gamma \mathbb{E}_{a' \sim \pi(s';\phi')}[\mathcal{D}_\mathbf{m}^lQ(s',a';\theta')]\big)\Big)^2 \Big],
\end{aligned}
\label{eq: consistent dropout bellman update}
\end{equation}
which means we apply the same mask matrix $\mathbf{m}$ to both sides of Bellman equation. Thus minimalist ensemble consistent Bellman update can eliminate the mismatch without destroying the diversity of value functions so that good value functions are learned and then a good policy can be derived. The diversity and ensemble properties of value functions hold due to the dropout operator. 
\begin{algorithm}[t]
	\caption{MEPG framework}
	\label{algo: mepg}
	\textbf{Initialize}: actor network $\pi$, critic network $Q$ with parameters $\phi, \theta$, target networks $\phi' \leftarrow \phi$, $\theta' \leftarrow \theta$, and replay buffer $\mathcal{B}$\\
	\textbf{Parameters}: $\text{num of samples } T, \text{probability of dropping } p, \eta, $ and $t=0$
	\begin{algorithmic}[1] 
		\STATE Reset the environment and receive initial state $s$
		\WHILE{$t < T$} 
		\STATE Select action a w.r.t. its policy network $\pi$ and receive reward $r$, new state $s'$
		\STATE Store transition tuple $(s, a, r, s')$ to $\mathcal{B}$
		\STATE Sample mini-batch of $N$ transitions $(s, a, r, s')$ from replay buffer $\mathcal{B}$
		\STATE Sample action $a' \sim \pi(s'; \phi')$ 
		\STATE Sample $\mathbf{m} \sim \text{Bernoulli}(1-p)$
		\STATE Compute target for the Q-function:
		\STATE $y \leftarrow r + \gamma \mathcal{D}_\mathbf{m}^{l}Q(s', a';\theta')$
		\STATE Update $\theta$ by one step gradient descent using:
		\STATE $\nabla_\theta J(\theta) =  N^{-1} \sum \nabla_\theta \frac{1}{2}\Big(y-\mathcal{D}_\mathbf{m}^{l}Q(s,a; \theta)\Big)^2$
		\STATE Update $\phi$ by one step of gradient ascent using Policy Gradient with learning rate $\epsilon$:
		\STATE $\phi \leftarrow \phi + \epsilon \nabla_\phi J(\phi)$
		\STATE Update target networks:
		\STATE $\theta' \leftarrow \eta \theta + (1-\eta) \theta'$, $\phi' \leftarrow \eta \phi + (1-\eta) \phi'$
		\STATE $t\leftarrow t+1$
		\STATE $s\leftarrow s'$	
		\ENDWHILE
	\end{algorithmic}
\end{algorithm}

\subsection{MEPG framework}
The MEPG framework is formulated by using the minimalist ensemble consistent Bellman update (equation (\ref{eq: consistent dropout bellman update})) in the policy evaluation phase and the conventional policy gradient methods in the policy improvement phase. At each learning step in MEPG framework, a $\mathcal{D}_\mathbf{m}^l$ operator is sampled, then policy evaluation phase proceeds with being acted upon by $\mathcal{D}_\mathbf{m}^l$. For policy improvement phase, the complete (i.e., ensemble) value function is used to train policy network. MEPG framework is summarized in Algorithm \ref{algo: mepg}. We apply the MEPG framework to the DDPG algorithm and SAC algorithm, and call the resulting algorithms ME-DDPG and ME-SAC respectively. 

In deterministic policy gradient style DRL algorithms, the policy can be updated by its value function \citep{dpg,ddpg}
\begin{equation}
\nabla_\phi J_{\text{DDPG}} ^ \pi (\phi) = \mathbb{E}_{s\sim P_\pi} [\nabla_a Q^\pi(s,a;\theta)|_{a=\pi(s;\phi)} \nabla_\phi \pi(s;\phi)],
\label{eq: dpg}
\end{equation}
where the actor $\pi$ and critic $Q$ are parameterized by $\phi$ and $\theta$ respectively. 
The target network is updated by $\theta' \leftarrow \eta \theta + (1-\eta) \theta'$ at each time step, where $\eta$ is a small constant. 

For the policy improvement phase in ME-DDPG, we utilize the original Deterministic Policy Gradient method (Equation (\ref{eq: dpg})) without being acted by $\mathcal{D}_\mathbf{m}^l$ operator. And for the policy evaluation phase, the ME-DDPG algorithm utilizes equation (\ref{eq: consistent dropout bellman update}). Besides, we take two tricks from TD3 algorithm \citep{td3}: target policy smoothing regularization and delayed policy updates. ME-DDPG algorithm is summarized in Appendix \ref{algo: me-ddpg}.  

The policy optimization objective of SAC \citep{sac} is
\begin{equation}
J^\pi_{\text{SAC}}(\phi) = \mathbb{E}_{s \sim \mathcal{B}} [\mathbb{E}_{a\sim \pi(\cdot \mid s, \phi)}[\alpha \log (\pi(a \mid s;\phi)) - Q(s,a; \theta)]].
\label{eq: sac policy objective}
\end{equation}
ME-SAC keeps the original policy update style in SAC. The soft Q-function of ME-SAC can be obtained by minimizing the soft Bellman residual following equation \ref{eq: consistent dropout bellman update},
\begin{equation}
\begin{aligned}
	J^Q_{\text{ME-SAC}}(\theta) = & \mathbb{E}_{(s,a) \sim \mathcal{B}} \Big[ \frac{1}{2} \Big( \mathcal{D}_\mathbf{m}^l Q(s,a;\theta)  \\
					&- \big(r(s,a) + \gamma \mathbb{E}_{s'\sim P}[\mathcal{D}_\mathbf{m}^l V(s'; \theta')] \big) \Big)^2\Big], \text{with}
\end{aligned}
\label{eq: sac Q objective}
\end{equation}
$$\mathcal{D}_\mathbf{m}^l V(s;\theta') = \mathbb{E}_{a\sim \pi(\cdot | s;\phi)} [ \mathcal{D}_\mathbf{m}^l Q(s,a;\theta') - \alpha \log \pi(a|s;\phi)].$$ The temperature parameter $\alpha$ can be given as a hyper-parameter or learned by minimizing 
\begin{equation}
J^{\alpha}_{\text{SAC}}(\alpha)  =  \mathbb{E}_{a\sim \pi^*} [- \alpha \log \pi^* (a|s; \alpha, \phi) -\alpha \mathcal{H}],
\label{eq: sac alpha objective}
\end{equation}
where $\mathcal{H}$ is a pre-given target entropy.

And for ME-SAC, we only use one critic with delayed policy updates. ME-SAC is summarized in Appendix \ref{algo: me-sac}. Note that any exploration technique can be coupled with MEPG (e.g. noise exploration in DDPG, entropy exploration in SAC).

\begin{figure*}[t]
	
	\centering
	\subfigure[Ant]{
		\includegraphics[width=1.7in]{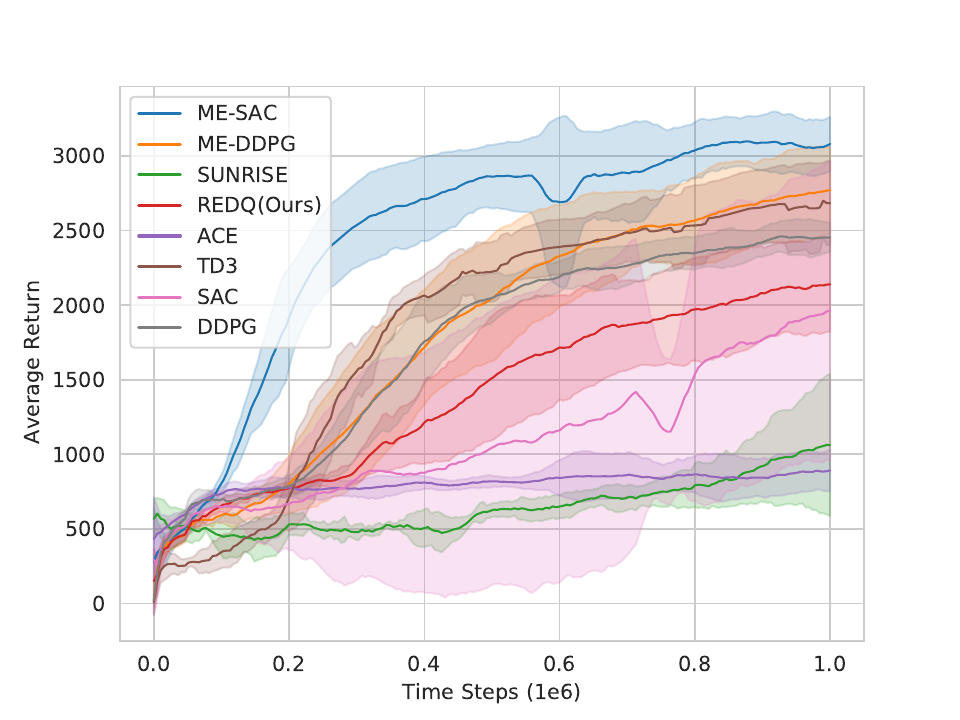}
	}
	\hspace{-0.3in}
	\subfigure[Walker2D]{
		\includegraphics[width=1.7in]{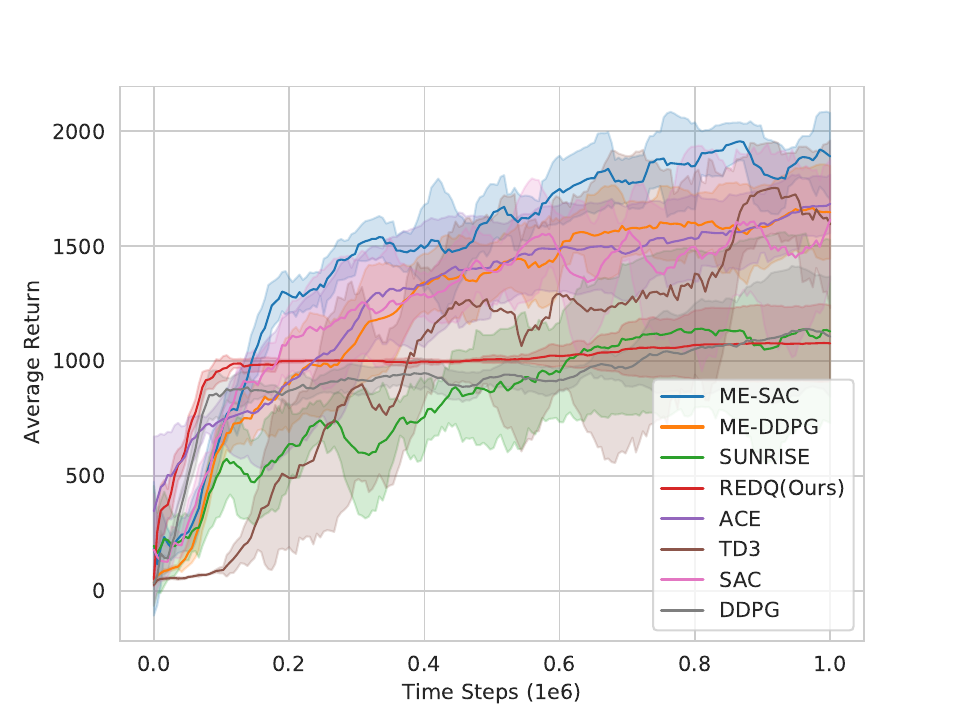}
	}
	\hspace{-0.3in}
	\subfigure[HalfCheetah]{
		\includegraphics[width=1.7in]{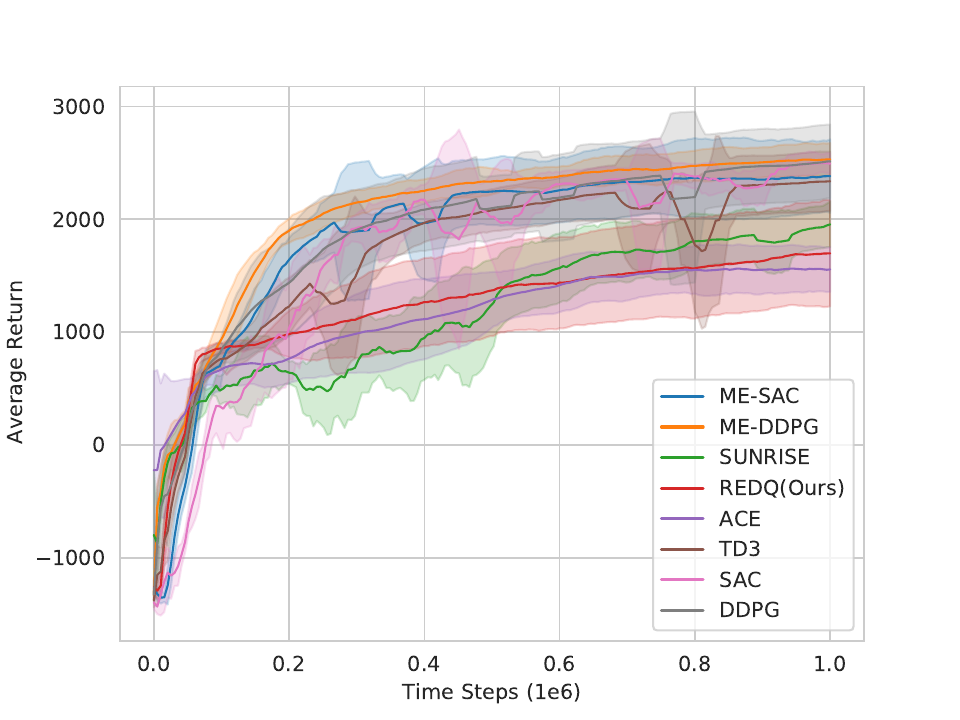}
	}   
	\hspace{-0.3in}
	\subfigure[Hopper]{
		\includegraphics[width=1.7in]{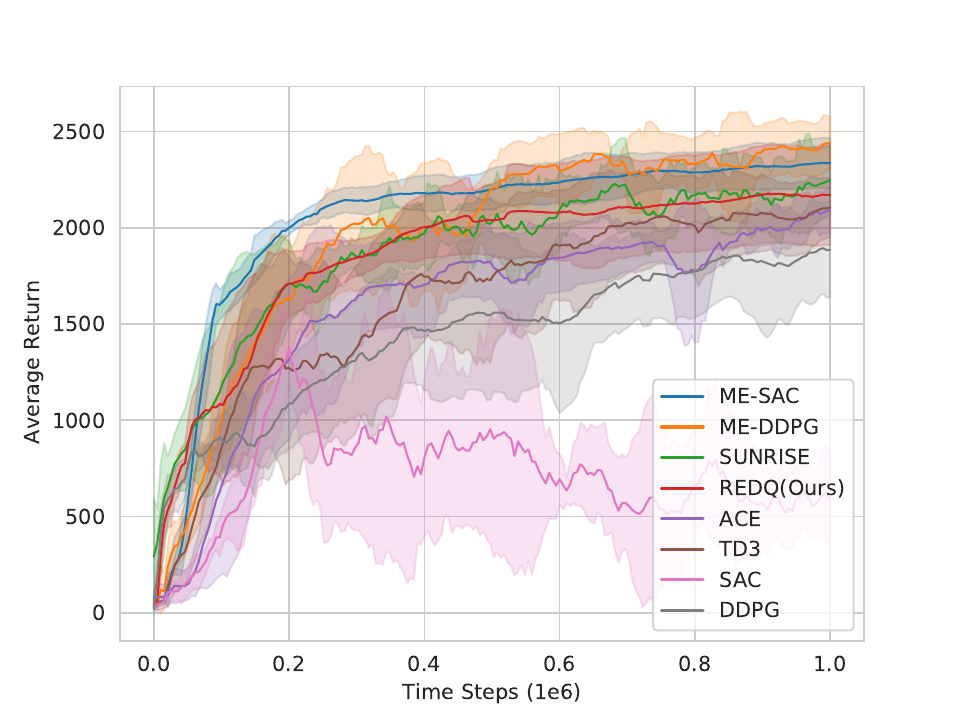}
	}
	\caption{Performance curves on gym PyBullet suite. The shaded area represents half a standard derivation of the average evaluation over 10 trials. For visual clarity, we use moving average smoothing curves with window size 10. Results show that the performance of the MEPG framework (ME-DDPG and ME-SAC) can match or outperform that of the tested algorithms.}
	\label{fig: performance learning process curve}
\end{figure*}

\subsection{Theoretical Analysis}\label{sec: theoretical analysis}
We show that dropout operator acting on value networks can be seen as a deep GP. Therefore, MEPG keeps consistency of the dropout mask acting on both sides of Bellman equation, then the two deep GPs induced by the Dropout operator acting on both sides of Bellman equation are kept synchronized. This consistency improves the stability of Bellman equation compared to random mask situations.

Let $\hat{Q}^\pi$ be the output of action value function (a neural network), and a loss function $\mathcal{F}(\cdots)$. Let $\mathbf{W}_i$ be the weight matrix of $M_i \times  M_{i-1}$ dimensions and the bias $ \mathbf{b}_i $  of layer $i$ of dimension $M_i$ for $i\in\{1,2,\cdots, L\}$. We define Bellman backup operator as 
\begin{equation}
\mathcal{T} Q^{\pi}(s,a; \theta) \overset{\text{def}}{=} r(s,a) + \gamma \mathbb{E}_{s' \sim P(\cdot|s,a), a'\sim \pi(s')}[Q^\pi(s',a'; \theta')]
\label{app eq: bellman backup}
\end{equation}
which is different from Bellman backup defined in tabular setting \citep{rl} due to importing auxiliary target network.
Let $Q^\pi_{\text{True}}$ be the fixed point of Bellman equation (\ref{eq: bellman equation Q}) w.r.t. policy $\pi$. In the DRL setting, the critic network, which characterizes action value, is used to find the fixed point of Bellman equation w.r.t. current policy $\pi$ through multiple Bellman update. This optimization method is a different paradigm from supervised learning. For convenient, we denote the input and output sets for critic as $\mathcal{X}$ and $\mathcal{Q}$ where $\mathcal{X} \subseteq \mathcal{S} \times \mathcal{A}$. For input $x_i$, the output of the action value function is $\hat{Q}_i$. We only discuss policy evaluation problems, i.e., how to approximate the true Q-function given policy $\pi$, thus we omit the $\pi$ in the following statement. For a more detailed analysis on the dropout operator behind deep learning, we would recommend the readers check \citep{gal2016dropout} for more analysis on this topic.  We often utilize modern optimization techniques like Adam \citep{kingma2015adam}, which utilizes the $L_2$ regularization term in the learning process. Thus the objective for policy evaluation in conventional DRL can be formulated as 
\begin{equation}
\begin{aligned}
\mathcal{L}_{\text{critic}} &= \mathbb{E}_{Q}\big[ \mathcal{F}(\mathcal{T}\hat{Q}, \hat{Q}) \big] + \lambda \sum_{i=1}^L \Big(\| \mathbf{W}_i\|_2^2 + \| \mathbf{b}_i\|_2^2\Big) \\
& \approx \frac{1}{N} \sum_{i=1}^N \mathcal{F}(\mathcal{T}\hat{Q}_i, \hat{Q}_i) + \lambda \sum_{i=1}^L \Big(\| \mathbf{W}_i\|_2^2 + \| \mathbf{b}_i\|_2^2\Big).
\end{aligned}
\label{app eq: loss function for critic}
\end{equation}
By minimizing Equation (\ref{app eq: loss function for critic}), we find the fixed point of Bellman equation, and then solve the policy evaluation problem. With the application of the dropout operator, the units of the neural network, are set to zero with probability $p$. Next we consider a deep Gaussian Process (GP) \citep{damianou2013deep}.
Now we are given a covariance function
\begin{equation}
\mathbf{C}(\mathbf{x},\mathbf{y}) = \int\limits_{\mathbf{w},b} d\mathbf{w}db \: p(\mathbf{w}) p(b) f(\mathbf{w}^\top \mathbf{x} + b) f(\mathbf{w}^\top \mathbf{y}+b),
\label{app eq: covariance}
\end{equation}
where $f$ is element-wise non-linear function. Equation (\ref{app eq: covariance}) is a valid covariance function. Assume layer $i$ have a parameter matrix $ \mathsf{W}_i $ with dimension $\mathsf{M}_i \times \mathsf{M}_{i-1}$ and we include all parameters in a set $\mathsf{\omega} = \{\mathsf{W}\}_{i=1}^L$. Let $p(\mathbf{w})$ in Equation (\ref{app eq: covariance}) is the distribution of each row of $ \mathsf{W}_i $. We assume that the dimension of the vector $\mathbf{m}_i$ for each GP layer is $\mathsf{M}_i$. Given some precision parameter $\varepsilon > 0$, the predictive probability of the Deep GP model is 
\begin{equation}
\begin{aligned}
&p(Q \mid \mathbf{x}, \mathcal{X,Q}) = \int \limits_{\omega}d\omega \: p(Q \mid x, \omega) p(\omega \mid \mathcal{X,Q})  \\
&p(Q \mid \mathbf{x}, \omega) = \mathcal{N} (Q; \hat{Q}(x; \omega), \varepsilon \mathbf{I}_D) \\
&\hat{Q}(\mathbf{x}; \omega) = \sqrt{\frac{1}{\mathsf{M}_L}} \mathsf{W}_L f\Big(\cdots \sqrt{\frac{1}{\mathsf{M}_2}} \mathsf{W}_2 f (\mathsf{W}_1 \mathbf{x} +\mathbf{m}_1) \cdots \Big).
\end{aligned}
\end{equation}

\begin{table}[htbp]
	\caption{The average of five maximum average returns over five trials of one million time steps for various algorithms. The maximum value for each task is bolded. + corresponds to a single standard deviation over trials. "HCheetah", "Walker" are shorthands for "HalfCheetah", and "Walker2D" respectively.}
	\label{table: performance table 4 results}
	\renewcommand\tabcolsep{3.0pt} 
	\begin{center}
        \begin{tabular}{ccccc}
        \toprule
        \textbf{Algorithm}&\textbf{Ant}&\textbf{HCheetah}&\textbf{Hopper}&\textbf{Walker}\\
        \midrule
        ME-SAC &\textbf{2907+284}&\textbf{3113+256}&2533+106&\textbf{1871+46}\\
        ME-DDPG &2841+262&2582+128&\textbf{2547+102}&1770+136\\
        REDQ{\footnotesize (Ours)}&2239+291&1757+420&2239+233&1093+167\\
        REDQ &2214+175&1327+294&2068+530&1009+3\\
        SUNRISE&1235+411&2018+194&2386+218&1270+324\\
        ACE &993+135&1635+192&2201+111&1730+111\\
        SAC &2009+854&2568+90&2318+132&1776+143\\
        TD3 &2758+227&2360+233&2190+181&1869+136\\
        DDPG &2533+103&2537+293&1969+221&1192+249\\

        \bottomrule
        \end{tabular}
	\end{center}
\end{table}

We utilize $ q(\omega) $ to approximate the intractable posterior $p(\omega; \mathcal{X,Q})$. Note that $q(\omega)$ is a distribution over matrices whose columns are randomly set to zero. We define $q(\omega)$ as 
\begin{equation}
\begin{aligned}
& \hat{\mathsf{G}}_i = \mathsf{G}_i  \odot \text{diag} ( [z_{i,j}]_{j=1}^{\mathsf{M}_i}), z_{i,j} \sim \text{Bernoulli} (1 - p_i),  \\
&\text{for} \: i \in \{1, \cdots, L\}, \: j\in\{1, \cdots, \mathsf{M}_{i-1}\}
\end{aligned}
\end{equation}
where $\mathsf{G}_i$ is a matrix as variational parameters. The variable $z_{i,j}=0$ means unit $j$ in layer $i-1$ being zero as an input to layer $i$, which recovers the dropout operator in neural networks. To learn the distribution $q(\omega)$, we minimize the KL divergence between $q(\omega) $ and $ p(\omega) $ of the full Deep GP
\begin{equation}
J_{\text{GP}} = - \int \limits_{\omega} d\omega \, q(\omega) \log p(\mathcal{Q} \mid \mathcal{X}, \omega) + \text{D}_{\text{KL}} (q(\omega) \| p(\omega)).
\label{app eq: deep gausian loss function}
\end{equation}
The first term in Equation (\ref{app eq: deep gausian loss function}) can be approximated by Monte Carlo method. We can approximate the second term in Equation (\ref{app eq: deep gausian loss function}), and obtain $\sum_{i=1}^L (\frac{p_i l^2}{2} \| \mathsf{G}_i \|_2^2 + \frac{l^2}{2} \| \mathbf{m_i}\|_2^2 ) $ with prior length scale $l$ (see section 4.2 in \citep{gal1506dropout_appendix}). Given the precision parameter $\varepsilon > 0$, the objective of deep GP can be formulated as 
\begin{equation}
\begin{aligned}
\mathcal{L}_{\text{GP}} \propto &\frac{1}{N\varepsilon } \sum_{i=1}^N - \log p(Q_i \mid \mathbf{x}_i ; \hat{\omega})  \\
&+ \frac{1}{N \varepsilon}\sum_{i=1}^L \Bigg( \frac{p_i l^2}{2} \|\mathsf{G}_i \|_2^2 +  \frac{l^2}{2} \|\mathbf{m}_i \|_2^2 \Bigg ).
\end{aligned}
\end{equation}
We can recover Equation (\ref{app eq: loss function for critic}) by setting $ \mathcal{F}(\mathcal{T}\hat{Q}, \hat{Q}) = - \log p(Q_i \mid x_i ; \hat{\omega}) $. Note that the sampled $\hat{\omega}$ leads to the realization of the Bernoulli distribution $z_{i,j}$, which is equivalent to the binary variable $z_{i,j}$ in the dropout operator.

The above analysis shows that the policy evaluation process in the MEPG framework maintains two synchronized deep GPs, because both sides of Bellman equation are acted by the same dropout mask $\mathcal{D}_\mathbf{m}^l$. The uncertainty introduced by the dropout operator arises from the inherent property if the model and explicitly exists in the model. Thus the diversity and ensemble properties of value functions hold. Besides, this consistency naturally improves the stability of Bellman equation compared to random mask situations.

\section{Experimental Results}


In this section, we answer the following questions: How good MEPG framework is superior to SOTA model-free and ensemble DRL algorithms? What is the contribution of each component to the algorithm? How do the training time cost and the number of parameters of our algorithm compared to other evaluated algorithms? How sensitive is MEPG to the fluctuations of hyper-parameters?

To evaluate our framework MEPG, we conduct experiments on open source PyBullet suite \citep{bullet3}, interfaced through Gym simulator \citep{gym}. We highlight the fact that the PyBullet suite is generally considered to be more challenging than the MuJoCo suite presented for continuous control tasks \citep{raffinSmoothExplorationRobotic2021}. Given the recent reproducibility discussions in DRL \citep{drl_matters,andrychowicz2021matters}, we strictly control all random seeds, and our results are reported over 5 trials unless otherwise stated, with the same setting and fair evaluation metrics. The experimental results are performed over multiple devices. More experimental results and more details can be found in the Supplementary Material.


\subsection{Evaluation}\label{sec: evaluation}
To evaluate the MEPG framework, we measure ME-DDPG and ME-SAC performance on PyBullet tasks compared with the SOTA model-free algorithm and recently proposed ensemble DRL algorithms such as ACE \citep{zhang2019ace}, SUNRISE \citep{sunrise} and REDQ \citep{redq}. For TD3, ACE, SUNRISE and REDQ, we use the authors' implementation with default hyper-parameters and keep the same hyper-parameters for DDPG and SAC. For a fair comparison, we replace one-time policy update every twenty times of critic optimization with one-time policy update every two-time of critic optimization in REDQ, namely REDQ(Ours). The empirical evaluation results show REDQ(Ours) is better than its original version. We run each task for one million time steps and perform one gradient step after each interaction step. Every $5,000$ time step, we execute an evaluation step over 10 episodes without any exploration operation in every algorithm.
Our performance comparison results are presented in Table \ref{table: performance table 4 results} and the learning curves are in Figure \ref{fig: performance learning process curve}. The results show that most of the time our proposed algorithm ME-DDPG and ME-SAC are better than state-of-the-art methods without any auxiliary tasks. For the Pendulum family of environments, all algorithms are equally good. Our framework is best in terms of learning speed and final performance for the Ant, Walker2D, and Hopper environments while consuming fewer computational resources. More experimental results and details are in the Supplementary Material.


\begin{table}[htbp]
	\caption{The average of the five maximum average returns over five trials of one million time steps for ablation studies. $\pm$ means adding or removing the corresponding component. The maximum value for each task is bolded. "MED", "MES", "MED-R" and "MES-R" are shorthands for "ME-DDPG", "ME-SAC", "ME-DDPG-R" and "ME-SAC-R" respectively.}
	\label{table: ablation for me-ddpg 4 env}
	\renewcommand\tabcolsep{3.0pt} 
	\begin{center}
		\begin{tabular}{ccccc}
			\toprule
			\textbf{Algorithm}&\textbf{Ant}&\textbf{HCheetah}&\textbf{Hopper}&\textbf{Walker}\\
			\midrule
MED &\textbf{2841.04} &\textbf{2582.32} &\textbf{2546.56} &\textbf{1770.11}\\
MED-DO &2001.56 &2522.97 &2326.2 &1774.61\\
MED-DU &2610.23 &2575.64 &2330.16 &1784.34\\
MED-TPS &2623.13 &2605.99 &2328.32 &1675.31\\
MED-R&2625.05 &2246.73 &2233.52 &1671.6\\
DDPG &2446.75 &2501.24 &1918.42 &1142.71\\

			\midrule
MES &\textbf{2906.98} &\textbf{3113.21}&\textbf{2532.98} &\textbf{1870.53}\\
MES-DU &2605.52 &2718.17 &2211.27 &1631.99\\
MES+FIXENT &818.03 &733.59 &1632.85 &843.01\\
MES-R &1530.76 &1905.75 &2112.46 &1651.07\\
SAC &2009.36 &2567.7 &2317.64 &1776.34\\
			\bottomrule
		\end{tabular}
	\end{center}
\end{table}

\subsection{Ablation Studies}\label{sec: ablation study}
We conduct ablation studies to demonstrate the contribution of each individual component. We show the ablation results for ME-DDPG and ME-SAC in Table \ref{table: ablation for me-ddpg 4 env}, where we compare the performance of removing or adding specific components from ME-DDPG or ME-SAC. Firstly, the effectiveness of the minimalist ensemble consistent Bellman update is investigated. We take the conventional Dropout operator acting on both sides of the Bellman equation, resulting in ME-DDPG-R and ME-SAC-R methods respectively. The empirical evaluations show that the proposed minimalist ensemble consistent Bellman update helps a lot and brings performance improvements. Secondly, we investigate three key components, i.e., dropout (DO), Target Policy Smoothing (TPS), and Delay Update (DU) in ME-DDPG. DO, DU and TPS help a lot. Thirdly, we perform a similar ablation analysis for the ME-SAC algorithm, where "FIX-ENT" means we adopt a fixed entropy coefficient $\alpha$. The ablation results of the ME-DDPG algorithm for DU, and DO are also applicable to the ME-SAC method. Here we find that automatic adjustment of entropy is extremely helpful. Additional experimental results and learning curves can be found in the Supplementary Material. 

%

\subsection{Run time and number of parameters}
We evaluate the run time of one million time steps of training for each tested RL algorithm. In addition, we also quantify the number of parameters of neural networks corresponding to the evaluated algorithms in different environments. For a fair comparison, we keep the same update steps in REDQ(M). Note that REDQ(M) is fairly faster than the original REDQ (see section \ref{sec: evaluation}). And we cancel the evaluation process in this process. We show the run time test results in Table \ref{table: run time} and parameters quantification in Table \ref{table: number of parameters}. The MEPG framework consumes the shortest time with the same skeleton algorithms. Unsurprisingly, We find our algorithm ME-DDPG and ME-SAC act favorably compared to other algorithms we tested in terms of wall-clock training time. Our MEPG framework not only runs in one-third to one-seventh of the time of ensemble learning methods, but even less than the training time of model-free DRL algorithms. All run time experiments are conducted on a single GeForce GTX 1070 GPU and an Intel Core i7-8700K CPU at 2.4GHZ. The number of parameters of the MEPG framework is between 14\% and 27\% of the number of parameters of the tested ensemble algorithms, but our algorithms perform better than any tested one.
\begin{table}[h]
	\caption{Run time comparison of training each RL algorithm.}
	\label{table: run time}
	\renewcommand\tabcolsep{3.0pt} 
	\begin{center}
		\begin{tabular}{ccccc}
			\toprule
			\textbf{Algorithm}&\textbf{Ant}&\textbf{HCheetah}&\textbf{Hopper}&\textbf{Walker}\\
			\midrule
			ME-DDPG &47m &44m &42m &45m\\
			TD3 &57m &55m &52m &55m\\
			DDPG &1h1m&58m&56m&58m\\
			\midrule
			ME-SAC &1h14m&1h13m&1h10m&1h12m\\
			SAC &1h17m&1h15m&1h12m&1h15m\\
			REDQ{\footnotesize(M)} &3h43m&3h41m&3h29m&3h35m\\
			SUNRISE &7h24m&7h15m&7h10m&7h16m\\
			\bottomrule
		\end{tabular}
	\end{center}
\end{table}

\begin{table}[h]
	\caption{The number of parameters is given in millions. For all tested environments, we utilize the same network architectures and the same hyper-parameters. The number of parameters differs because different environments have various state and action dimensions, which results in different input and output dimensions of the neural network.}
	\label{table: number of parameters}
	\renewcommand\tabcolsep{3.0pt}
		\begin{center}
	\begin{tabular}{ccccc}
		\toprule
		\textbf{Algorithm}&\textbf{Ant}&\textbf{HCheetah}&\textbf{Hopper}&\textbf{Walker}\\
		\midrule
	    ME-SAC &0.226M &0.223M &0.212M &0.220M\\
		ME-DDPG &0.302M &0.297M &0.283M &0.293M\\
		TD3 &0.453M &0.446M &0.425M &0.440M\\
		SAC &0.377M &0.372M &0.354M &0.367M\\
		REDQ &1.586M &1.564M &1.489M &1.543M\\
		SUNRISE &1.132M &1.117M &1.063M &1.101M\\
		ACE &1.614M &1.597M &1.532M &1.577M \\
		\bottomrule
	\end{tabular}
	\end{center}
\end{table}

\begin{figure}[t]
	
	\centering
	\subfigure[ME-DDPG]{
		\includegraphics[width=1.65in]{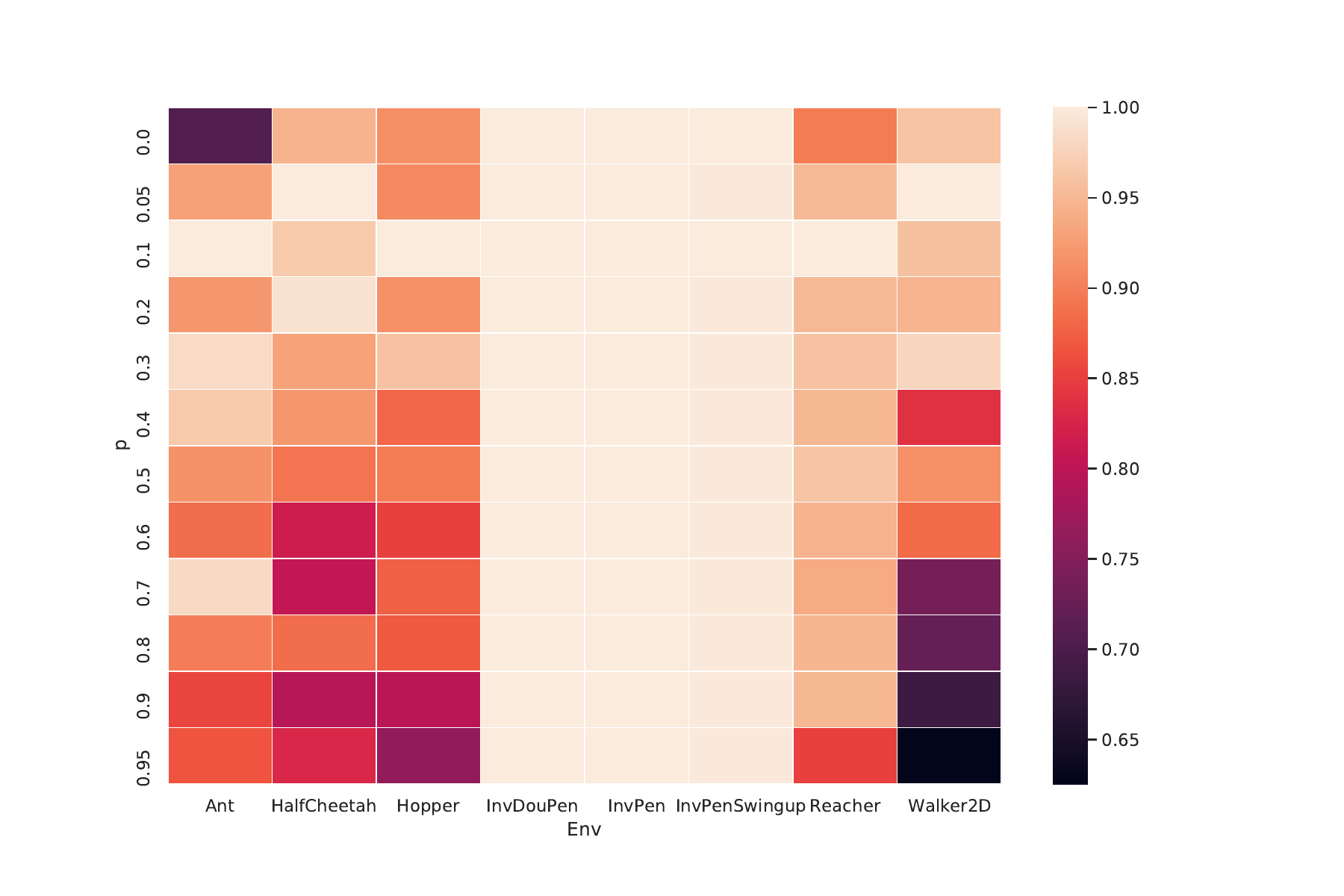}
	}
	\hspace{-0.3in}
	\subfigure[ME-SAC]{
		\includegraphics[width=1.65in]{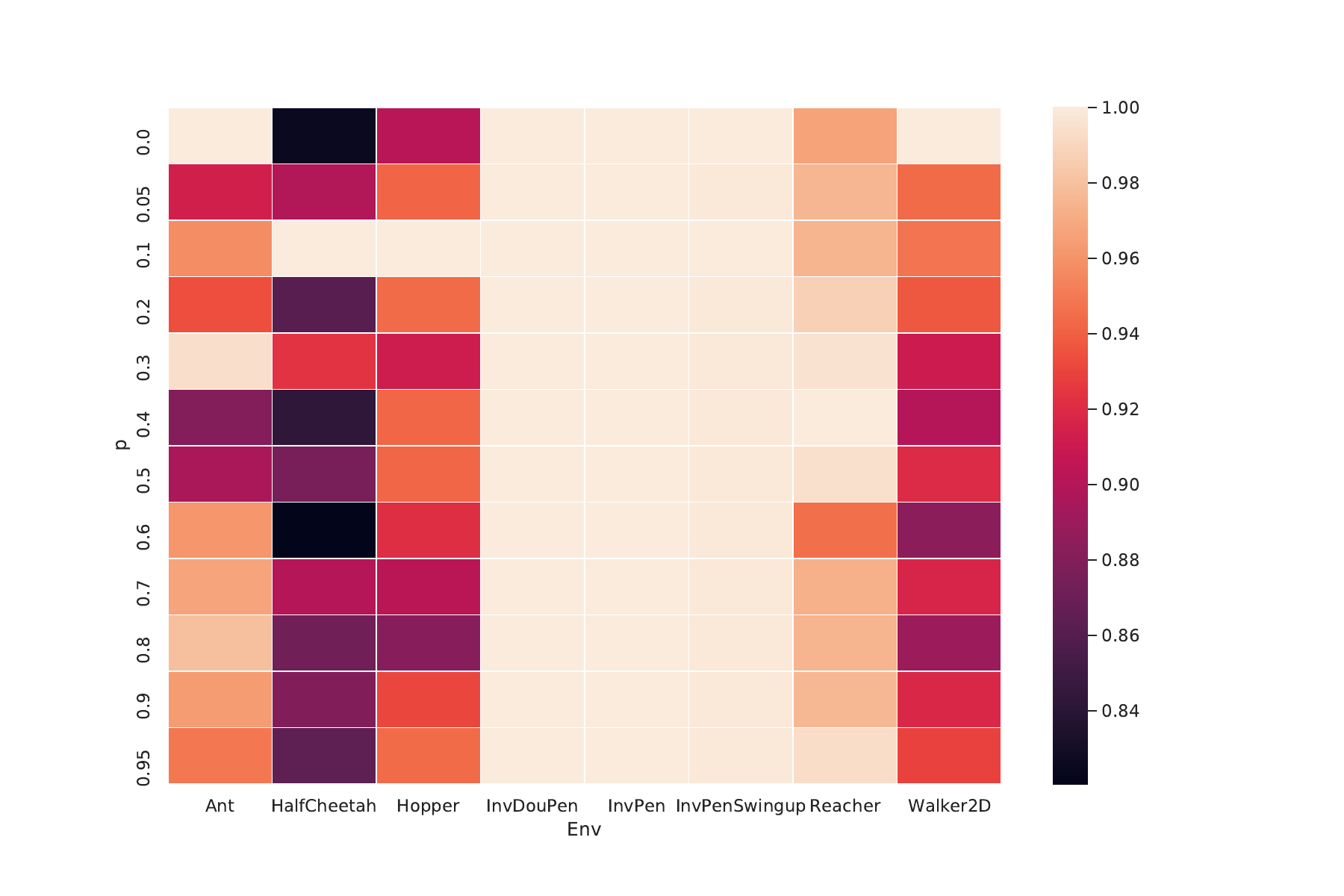}
	}
	\caption{Performance of ME-SAC and ME-DDPG algorithms given different p-values in different environments. Each cell represents the average of the top five maximum average returns in a run over five trials of one million time steps for various p-value. For most environments and p-values, both algorithms are able to learn successfully. Besides, the difference in performance between the different p-values is not very significant. Relatively speaking, smaller p-values give better performance. The performance of both algorithms performs best when $p=0.1$. }
	\label{fig: sensitive heat map}
\end{figure}

\subsection{Hyper-parameter Sensitivity}
The MEPG framework only introduces one hyper-parameter $p$, which represents the probability of setting a neuron of the neural network to zero. We take eleven $p$ values in interval $[0, 1]$. For visual simplicity, we normalize the data by $\text{Ret}^{\text{env}}_p = \text{Ret}^{\text{env}}_p / \max_p  \{\text{Ret}^{\text{env}}_p \}$, where $\text{Ret}^{\text{env}}_p$ means the average of top five maximum average returns in a run over five trials of one million time steps for various p-value on \texttt{env} environment. We show the results in Figure \ref{fig: sensitive heat map}. Each cell represents the result of a different p-value in various environments. The experimental results show that even large p-values, i.e., implicitly integrating enough models, do not cause the learning process to fail. Overall, the difference in performance between the different p-values is not very significant. For the Pendulum family of environments, the performance of the MEPG framework induced by various $p$ values is equally good. Relatively speaking, a smaller p-value gives better performance. Our MEPG framework is not extremely hyper-parameter sensitive. Therefore, we set $p=0.1$ as the default hyper-parameter setting for the ME-DDPG and ME-SAC algorithms.

\section{Conclusion}
Ensemble RL is an important class of methods for generalizable RL. However, it raises computational resource consumption and introduces more hyper-parameters. To apply ensemble deep RL algorithms to the real world, we need to solve the tricky problems mentioned in section \ref{sec: introduction}. Thus, we propose a novel ensemble RL framework, called \underline{M}inimalist \underline{E}nsemble \underline{P}olicy \underline{G}radient (MEPG). The ensemble property of MEPG is induced by a minimalist ensemble consistent Bellman update that utilizes a modified dropout operator. We show that the policy evaluation in the MEPG framework theoretically maintains two synchronized deep GPs, resulting in the stability of the Bellman equation. Next, we verify the effectiveness of MEPG in the PyBullet control suite. The experimental evaluations show that the performance of ensemble DRL algorithms can be easily outperformed by MEPG. Besides superior performance, MEPG has tremendous advantages in terms of run time and the number of parameters compared to the tested model-free and ensemble DRL algorithms. The core technique we adopt is the minimalist ensemble consistent Bellman update, wherein the dropout operator is a popular method in deep learning. MEPG can be used in real-world applications such as autonomous driving\citep{filos2020can}, where the computational burden is huge, and the task is time-critical. For the bigger picture of the RL community, we highlight that the potential of neural networks and relevant techniques may not have been fully exploited, as exemplified by our MEPG framework. For future work, we can explore the connections between techniques often used in deep learning and the problem structure of RL itself, such as the regularization of neural networks (e.g., layer normalization), and the properties that the features learned by the policy and value networks should have.

\bibliography{aaai22}

\clearpage

\section*{Supplementary Material}
\section{Missing Algorithms}
We give a detailed description about ME-SAC in Algorithm \ref{algo: me-sac} and ME-DDPG in Algorithm \ref{algo: me-ddpg} respectively.
\begin{algorithm}[htbp]
	\caption{ME-SAC}
	\label{algo: me-sac}
	\textbf{Initialize}: actor network $\pi$ and critic network $Q$ with parameters $\phi, \theta$, target networks $\theta' \leftarrow \theta$ \\
	\textbf{Initialize}: Replay buffer $\mathcal{B}$ 
	\textbf{Parameters}: total steps $T$, $p$, $t=0$, target entropy $\mathcal{H}$ and dropout probability $p$.
	\begin{algorithmic}[1] 
		\STATE Reset the environment and receive initial state $s$
		\WHILE{$t < T$} 
		\STATE Select action $a \sim \pi(\cdot \mid s;\phi)$, and receive reward $r$, new state $s'$
		\STATE Store transition tuple $(s, a, r, s')$ to $\mathcal{B}$
		\STATE Sample mini-batch of $N$ transitions $(s, a, r, s')$ from $\mathcal{B}$
		\STATE Sample $\mathbf{m} \sim \text{Bernoulli}(1-p)$
		\STATE Compute target for the Q-function:
		\STATE $y \leftarrow r + \gamma \mathcal{D}_\mathbf{m}^{l}Q(s', \tilde{a'};\theta') - \alpha \log \pi(\tilde{a}' \mid s' ;\phi), \; \tilde{a}' \sim \pi(\cdot \mid s'; \phi)$
		\STATE Update $\theta$ by one step of gradient descent using:
		\STATE $\nabla_\theta J(\theta) = \nabla_\theta N^{-1} \sum \Big( y-\mathcal{D}_\mathbf{m}^{l}Q_{\theta}(s,a)\Big)^2$
		\IF{$t$ mod $d$}
		\STATE Update $\phi$ by one step of gradient ascent using:
		\STATE $\nabla_\phi J(\phi) = N^{-1}\sum\nabla_a \Big(Q(s, \tilde{a};\theta) - \alpha \log \pi(\tilde{a} \mid s ;\phi)\Big), \;  \tilde{a} \sim \pi(\cdot \mid s; \phi)$
		\STATE Update $\alpha$ by one step gradient descent using:
		\STATE $\nabla_\alpha J(\alpha) =  N^{-1} \sum \nabla_\alpha \Big( - \alpha \log \pi (a \mid s; \alpha, \phi) - \alpha \mathcal{H} \Big)$
		\STATE Update target network:
		\STATE $\theta' \leftarrow \eta \theta + (1-\eta) \theta'$
		\ENDIF
		\STATE $t\leftarrow t+1$
		\STATE $s\leftarrow s'$	
		\ENDWHILE
	\end{algorithmic}
\end{algorithm}

\begin{algorithm}[t]
	\caption{ME-DDPG}
	\label{algo: me-ddpg}
	\textbf{Initialize}: actor network $\pi$, critic network $Q$ with parameters $\phi, \theta$, target networks $\phi' \leftarrow \phi$, $\theta' \leftarrow \theta$, and replay buffer $\mathcal{B}$\\
	\textbf{Parameters}: $T, p, \eta, d,$ and $t=0$
	\begin{algorithmic}[1] 
		\STATE Reset the environment and receive initial state $s$
		\WHILE{$t < T$} 
		\STATE Select action with noise $a = \pi(s;\phi) + \epsilon, \epsilon \sim \mathcal{N}(0, \sigma^{2}) $, and receive reward $r$, new state $s'$
		\STATE Store transition tuple $(s, a, r, s')$ to $\mathcal{B}$
		\STATE Sample mini-batch of $N$ transitions $(s, a, r, s')$ from $\mathcal{B}$
		\STATE $\tilde{a} \leftarrow \pi(s';\phi') + \epsilon$, $\epsilon \sim \text{clip}(\mathcal{N}(0, \tilde{\sigma}^2), -c, c)$
		\STATE Sample $\mathbf{m} \sim \text{Bernoulli}(1-p)$
		\STATE Compute target for the Q-function:
		\STATE $y \leftarrow r + \gamma \mathcal{D}_\mathbf{m}^{l}Q(s', \tilde{a};\theta')$
		\STATE Update $\theta$ by one step gradient descent using:
		\STATE $\nabla_\theta J(\theta) =  N^{-1} \sum \nabla_\theta \frac{1}{2}\Big(y-\mathcal{D}_\mathbf{m}^{l}Q_{\theta}(s,a)\Big)^2$
		\IF{$t$ mod $d$}
		\STATE Update $\phi$ by one step of gradient ascent using the Deterministic Policy Gradient:
		\STATE $\nabla_\phi J(\phi) =N^{-1} \sum\nabla_aQ(s, a;\theta)|_{a=\pi(s;\phi)} \nabla_\phi\pi(s;\phi)$
		\STATE Update target networks:
		\STATE $\theta' \leftarrow \eta \theta + (1-\eta) \theta'$, $\phi' \leftarrow \eta \phi + (1-\eta) \phi'$
		\ENDIF
		\STATE $t\leftarrow t+1$
		\STATE $s\leftarrow s'$	
		\ENDWHILE
	\end{algorithmic}
\end{algorithm}

\section{More Experimental Results and Details}
We mostly report experimental results on four environments in the body due to space limitations. In this section, we provide more experimental results and details on eight environments. Note that all the experimental results are conducted on the PyBullet suite. We highlight the fact that the PyBullet suite is usually considered harder to train than MuJoCo suite \citep{raffinSmoothExplorationRobotic2021} as we discussed. Besides, we strictly control all random seeds, and our results are reported over 5 trials unless otherwise stated, with the same setting and fair evaluation metrics. 

\textbf{Evaluation}. We provide empirical evaluation results on eight environments in Table \ref{app table: performance table 8 results}.

\begin{figure*}[ht]
	
	\centering
	\subfigure[Ant]{
		\includegraphics[width=1.8in]{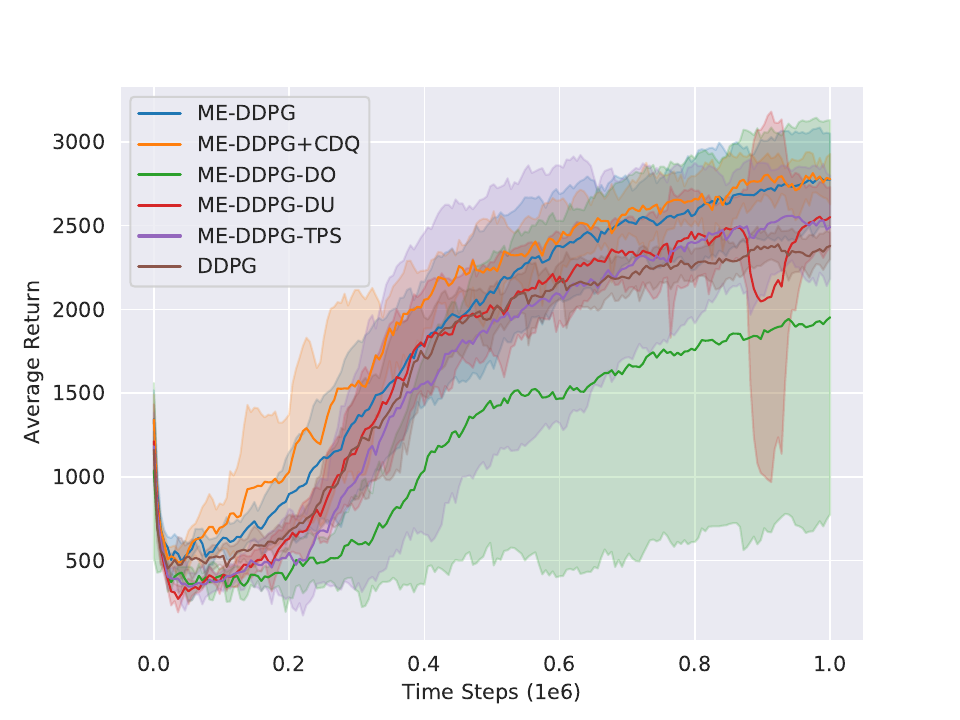}
	}
	\hspace{-0.3in}
	\subfigure[HalfCheetah]{
		\includegraphics[width=1.8in]{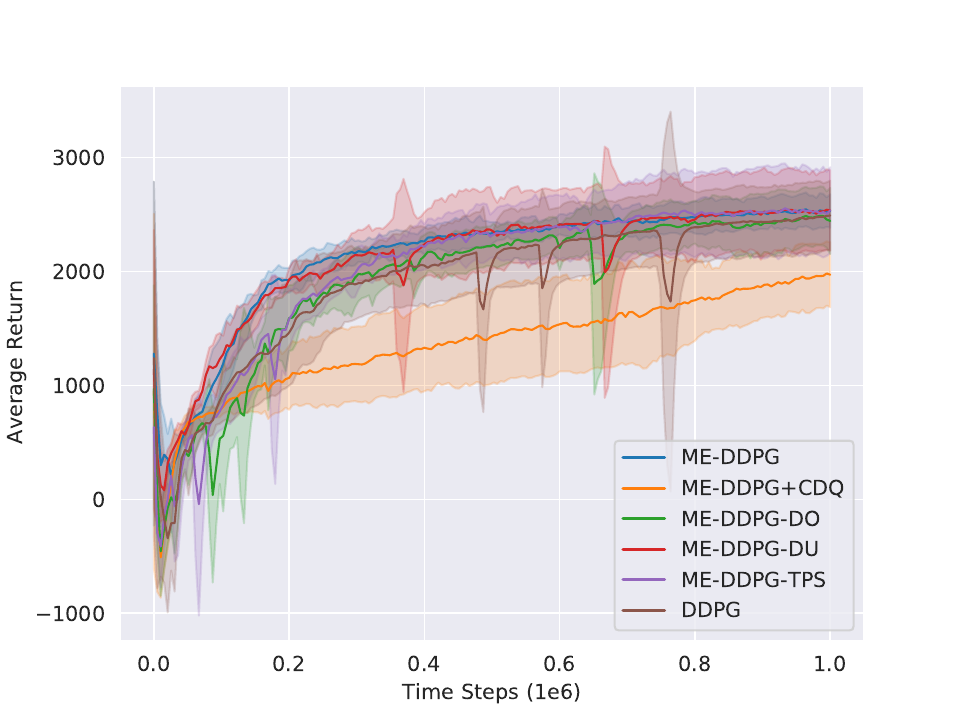}
	}   
	\hspace{-0.3in}
	\subfigure[Hopper]{
		\includegraphics[width=1.8in]{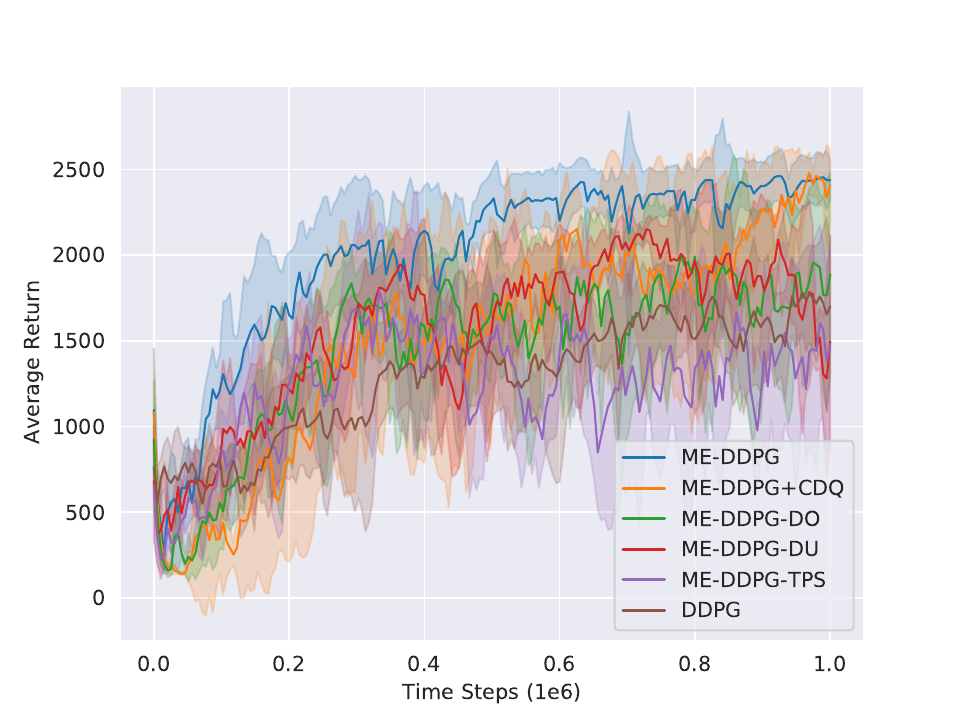}
	}
	\hspace{-0.3in}
	\subfigure[InvertedDoublePendulum]{
		\includegraphics[width=1.8in]{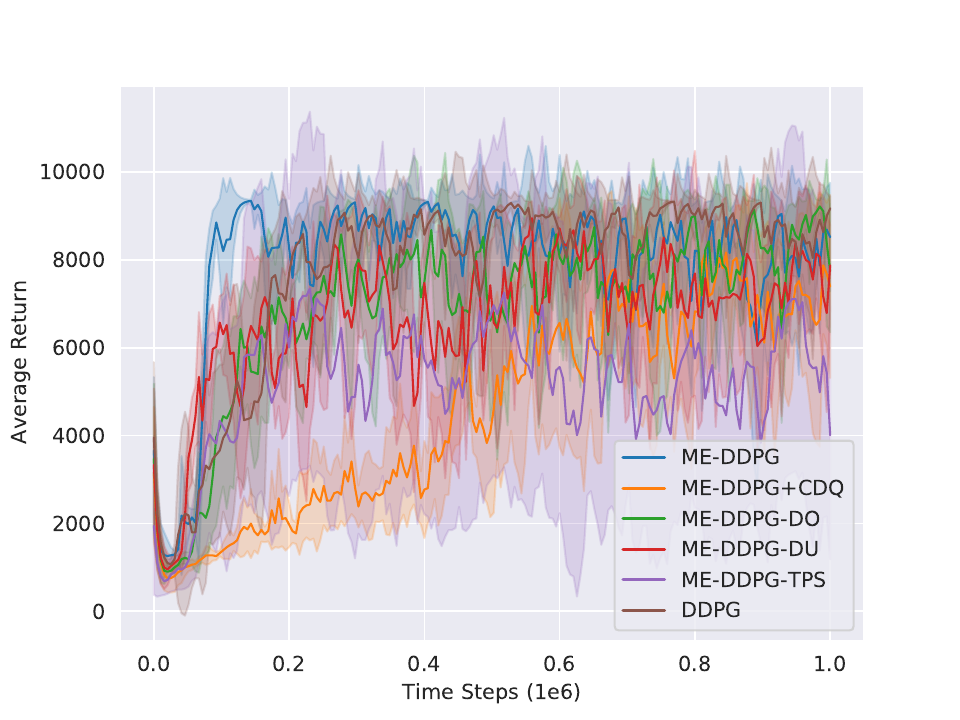}
	}
	\hspace{-0.3in}
	\subfigure[InvertedPendulum]{
		\includegraphics[width=1.8in]{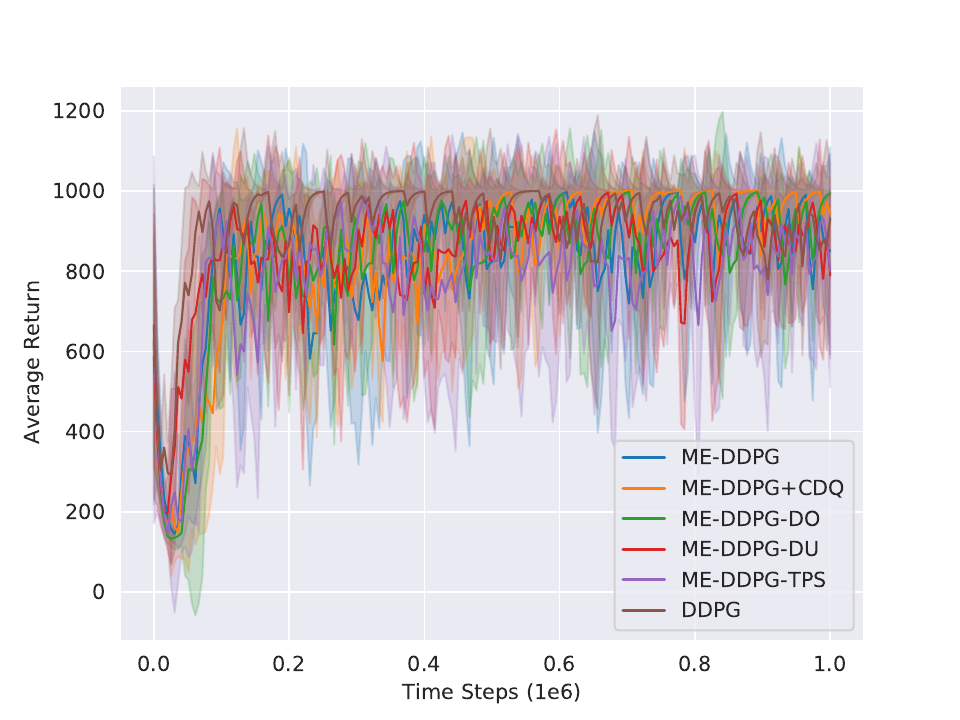}
	}
	\hspace{-0.3in}
	\subfigure[InvertedPendulumSwingup]{
		\includegraphics[width=1.8in]{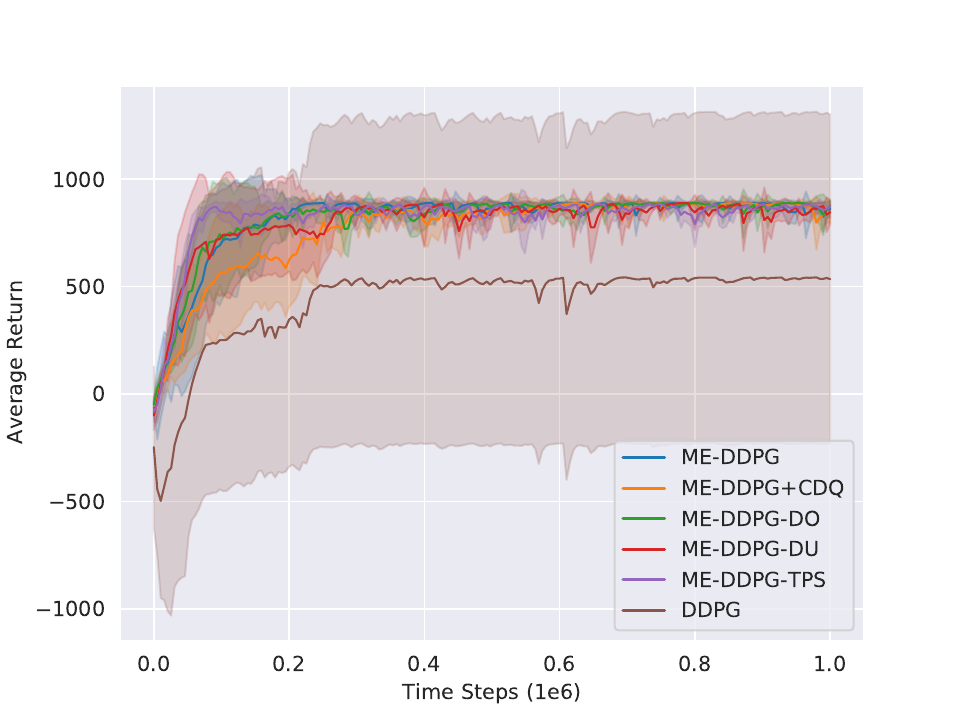}
	}   
	\hspace{-0.3in}
	\subfigure[Reacher]{
		\includegraphics[width=1.8in]{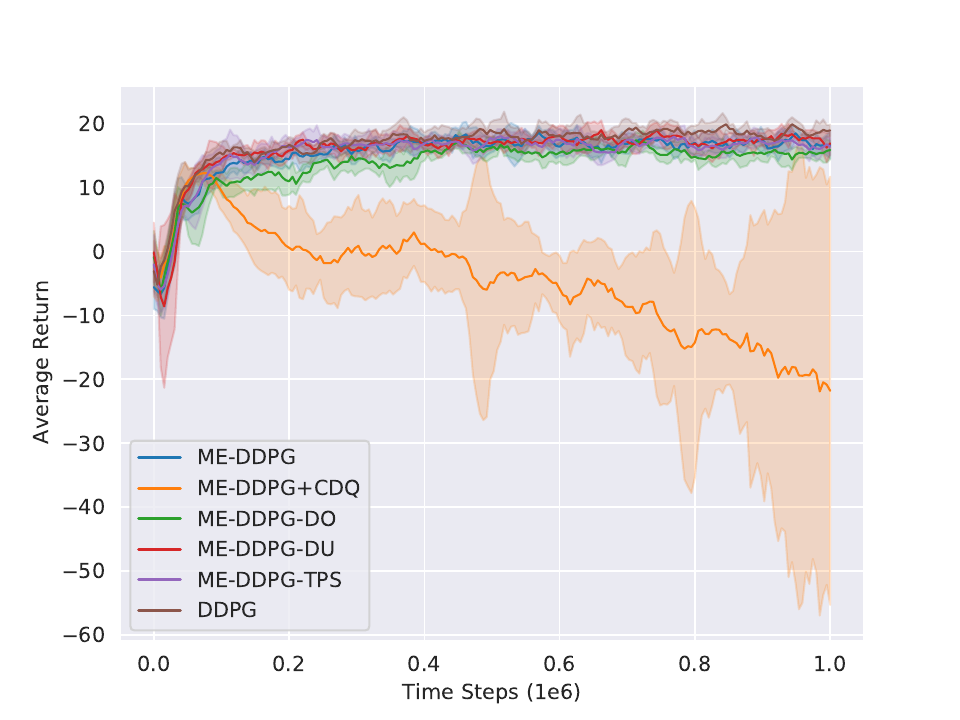}
	}
	\hspace{-0.3in}
	\subfigure[Walker2D]{
		\includegraphics[width=1.8in]{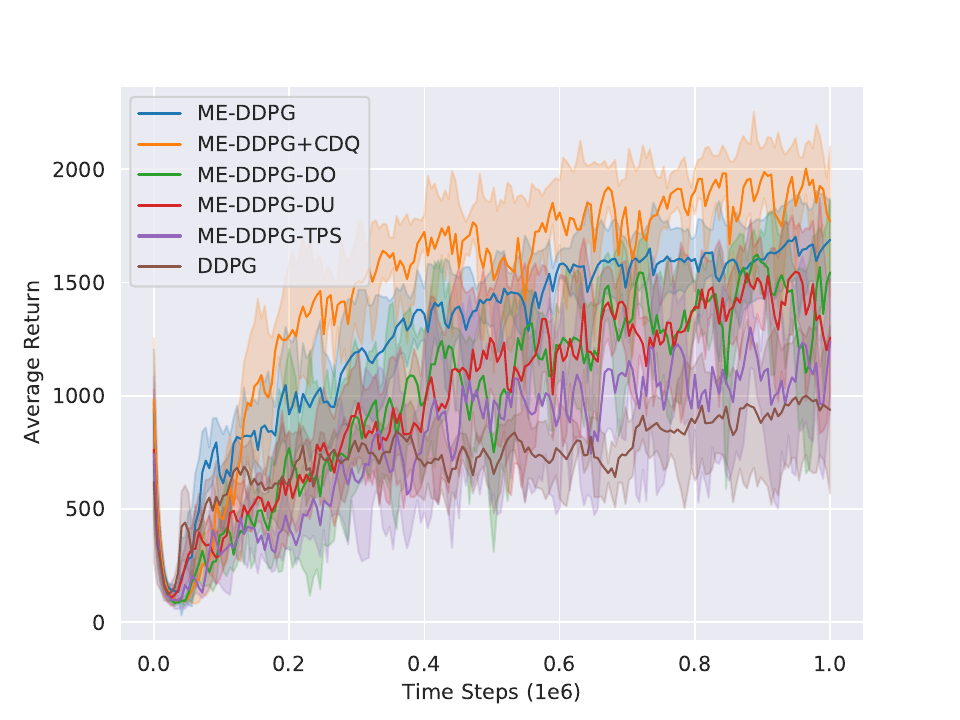}
	}
	\caption{Performance curves on gym PyBullet suite. The shaded area shows a half standard derivation.}
	\label{app fig: ablation for mepg-ddpg 8 env}
\end{figure*}

\begin{figure*}[ht]
	
	\centering
	\subfigure[Ant]{
		\includegraphics[width=1.8in]{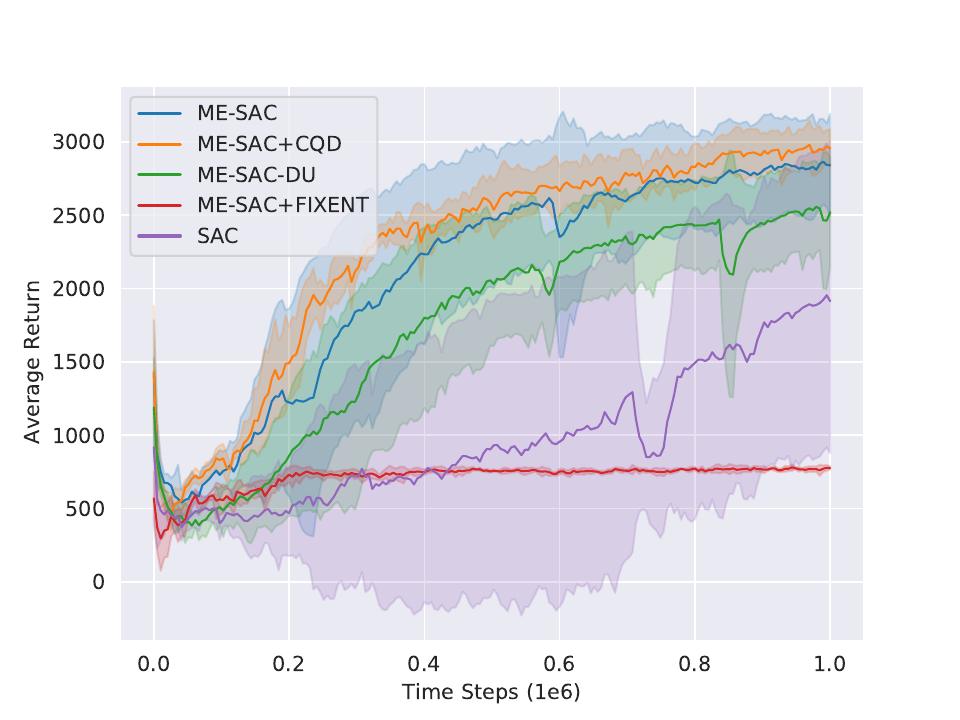}
	}
	\hspace{-0.3in}
	\subfigure[HalfCheetah]{
		\includegraphics[width=1.8in]{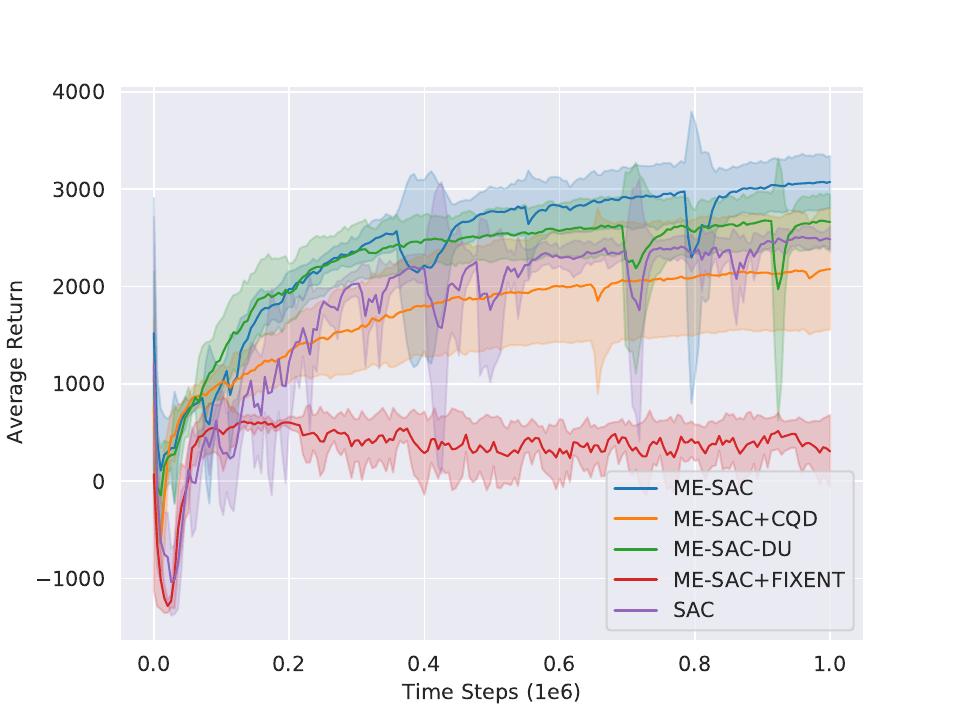}
	}   
	\hspace{-0.3in}
	\subfigure[Hopper]{
		\includegraphics[width=1.8in]{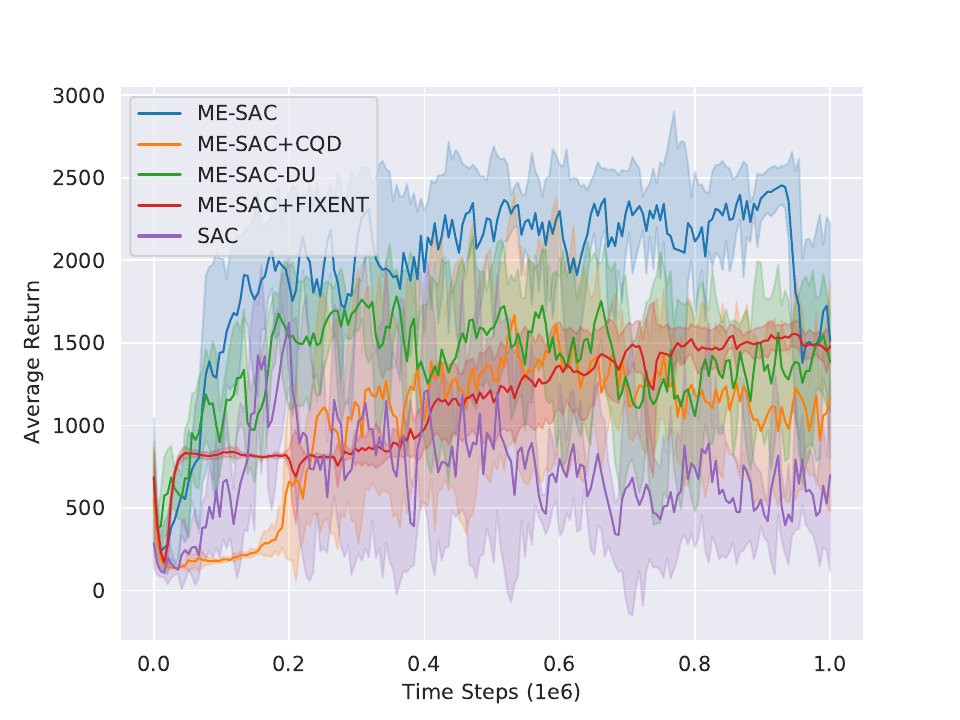}
	}
	\hspace{-0.3in}
	\subfigure[InvertedDoublePendulum]{
		\includegraphics[width=1.8in]{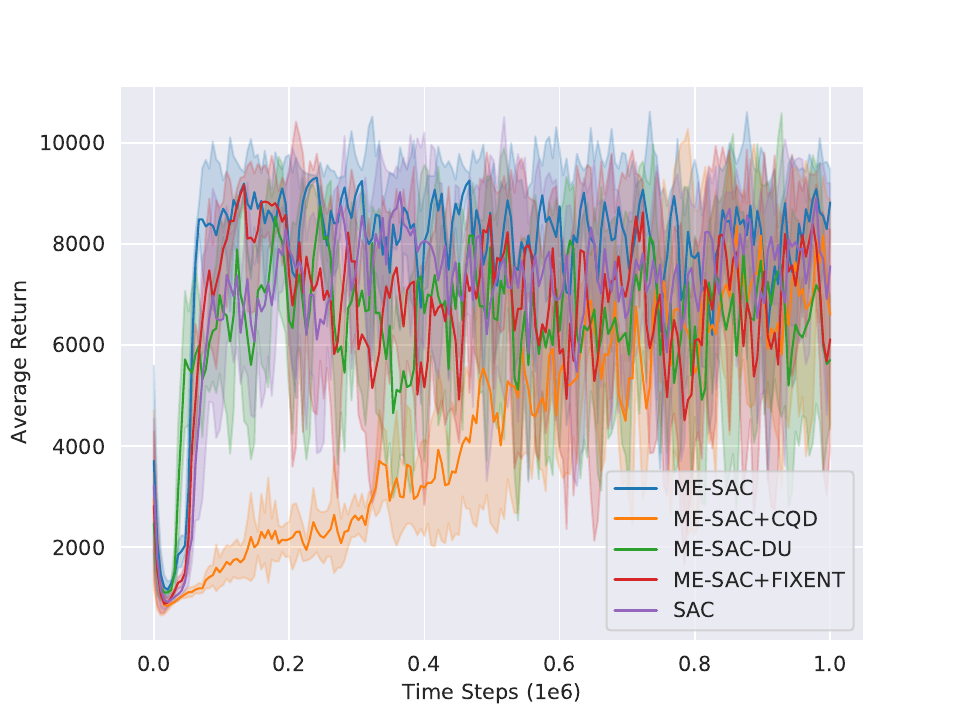}
	}
	\hspace{-0.3in}
	\subfigure[InvertedPendulum]{
		\includegraphics[width=1.8in]{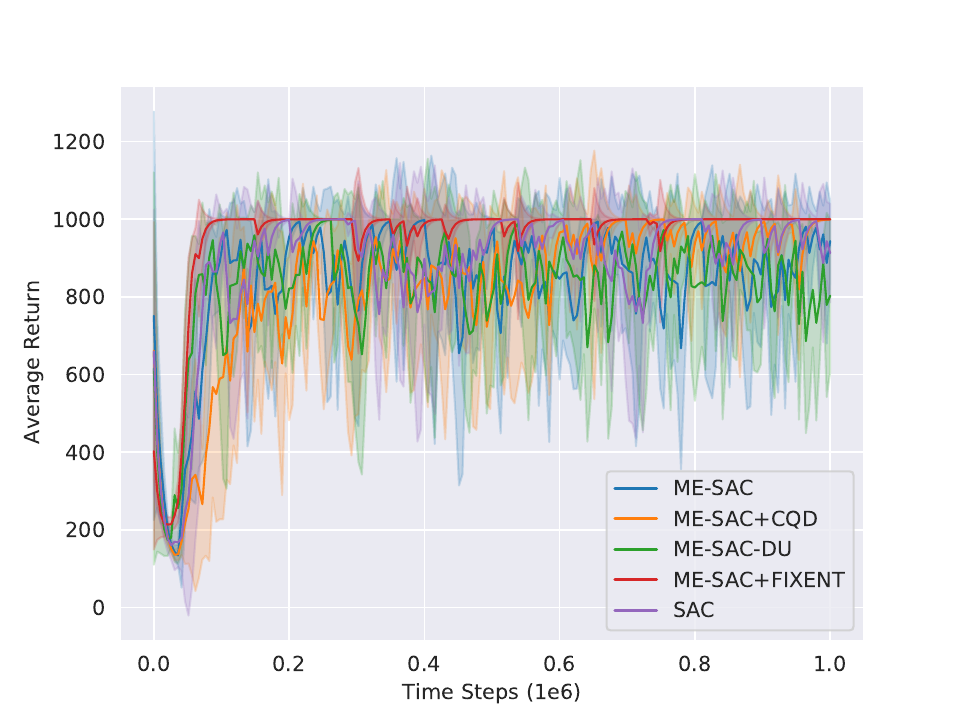}
	}
	\hspace{-0.3in}
	\subfigure[InvertedPendulumSwingup]{
		\includegraphics[width=1.8in]{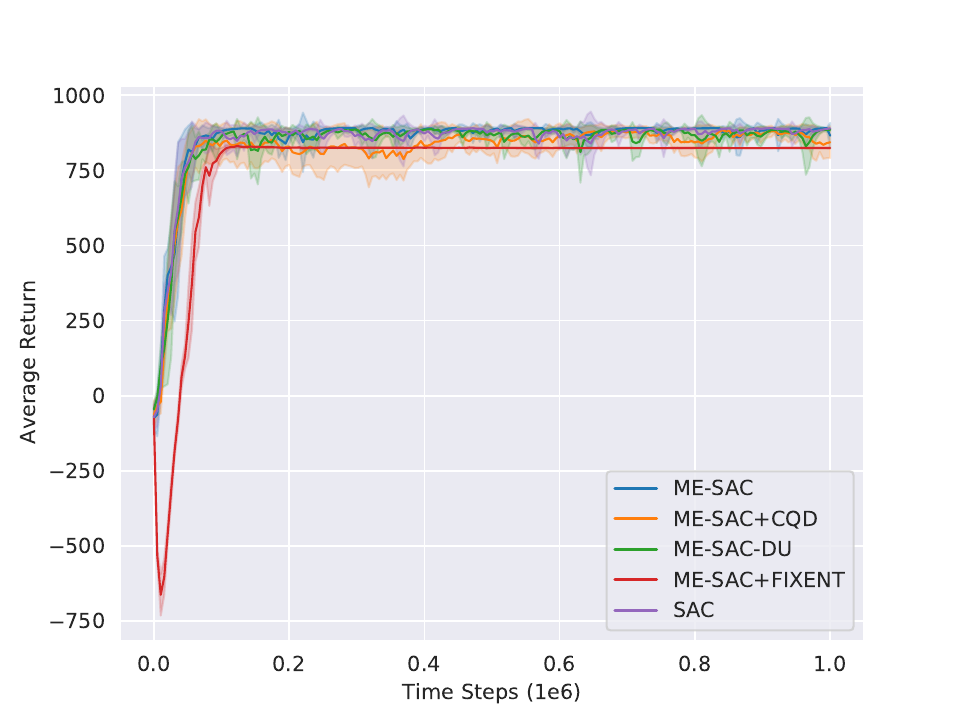}
	}   
	\hspace{-0.3in}
	\subfigure[Reacher]{
		\includegraphics[width=1.8in]{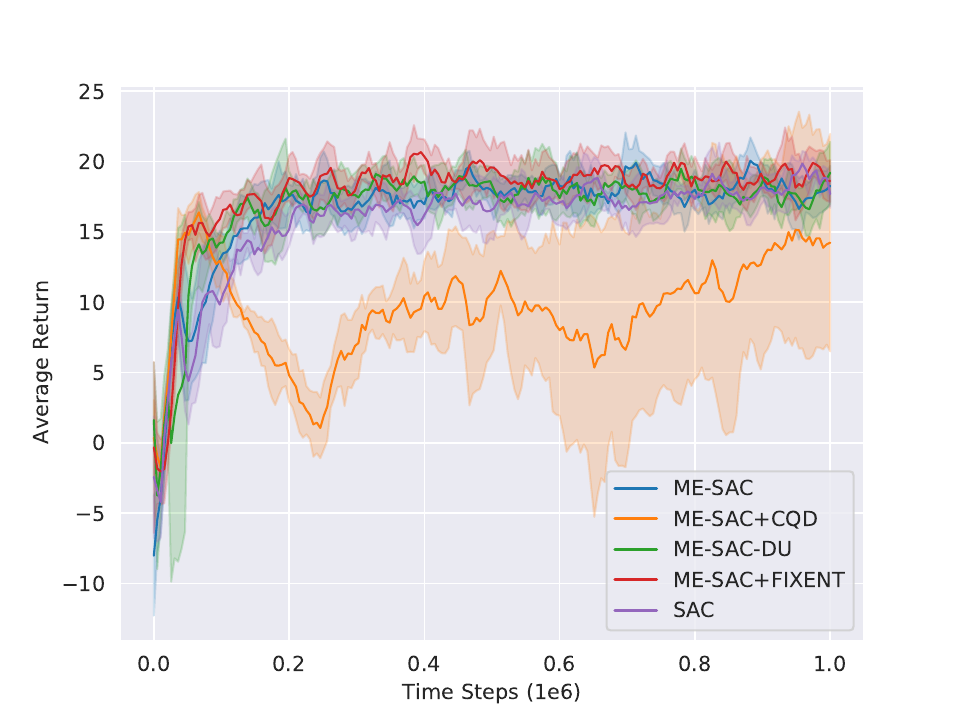}
	}
	\hspace{-0.3in}
	\subfigure[Walker2D]{
		\includegraphics[width=1.8in]{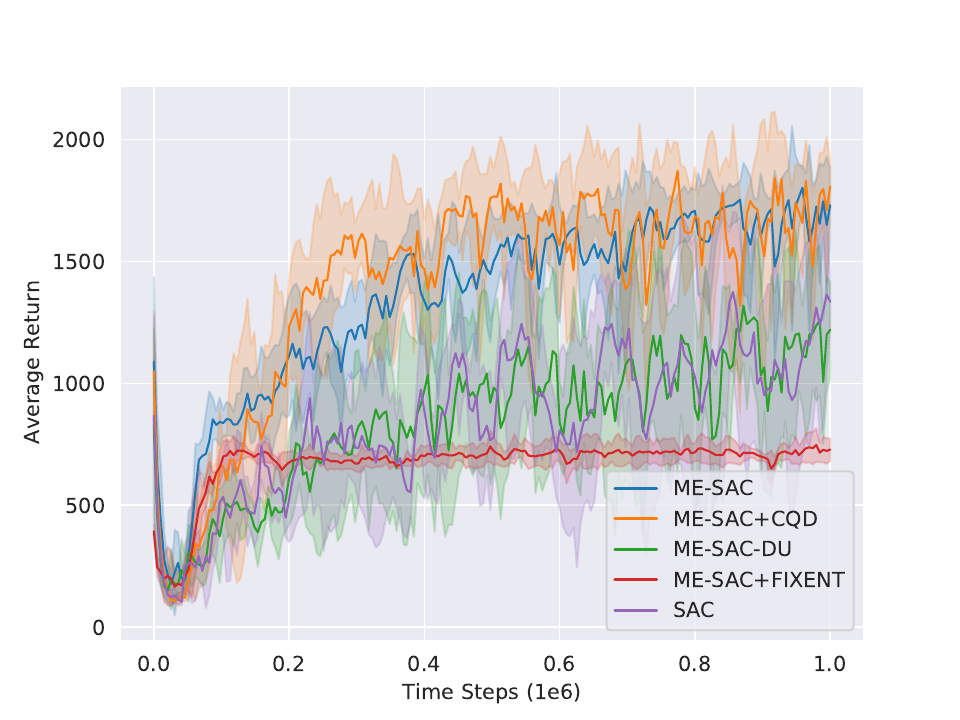}
	}
	\caption{Performance curves on gym PyBullet suite. The shaded area shows a half standard derivation.}
	\label{app fig: ablation for me-sac}
\end{figure*}

\textbf{Ablation}. We provide the learning curves for ablation analysis for ME-DDPG and ME-SAC in Figure \ref{app fig: ablation for mepg-ddpg 8 env} and Figure \ref{app fig: ablation for me-sac} respectively. And We also compute the top five maximum average returns of one million time steps for ablation analysis, which are shown in Table \ref{app table: ablation for me-ddpg} and Table \ref{app table: ablation for me-sac}. 

\textbf{Run time}. We show the run time statistics in Table \ref{app table: run time}.

\textbf{Number of parameters}. We show the number of parameters statistics in Table \ref{app table: number of parameters}.

\begin{figure*}[htbp]
	
	\centering
	\subfigure[Ant]{
		\includegraphics[width=1.8in]{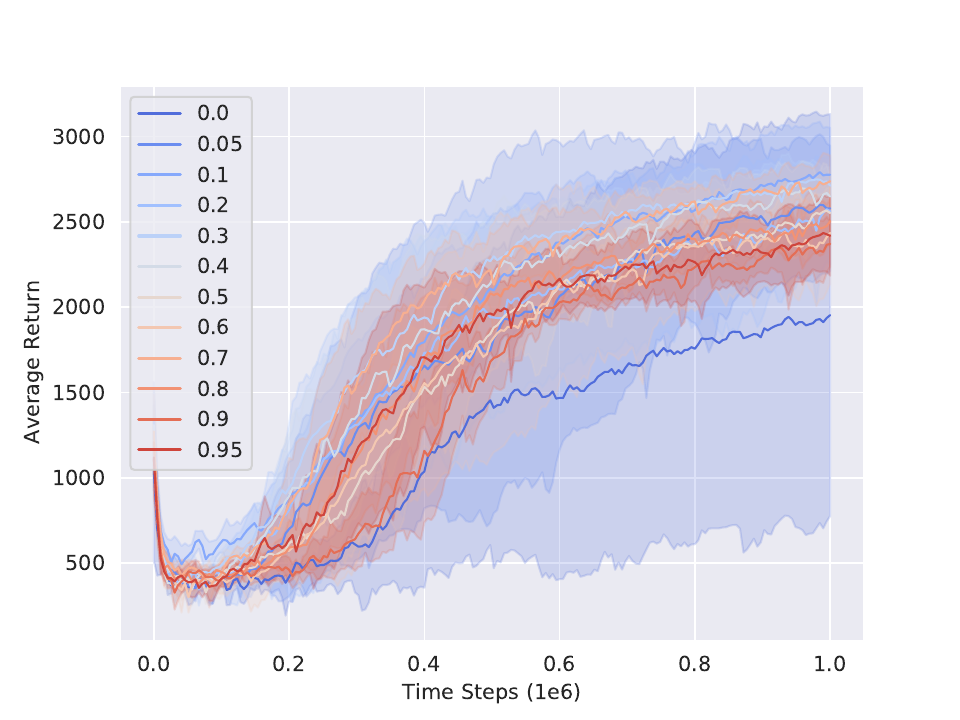}
	}
	\hspace{-0.3in}
	\subfigure[HalfCheetah]{
		\includegraphics[width=1.8in]{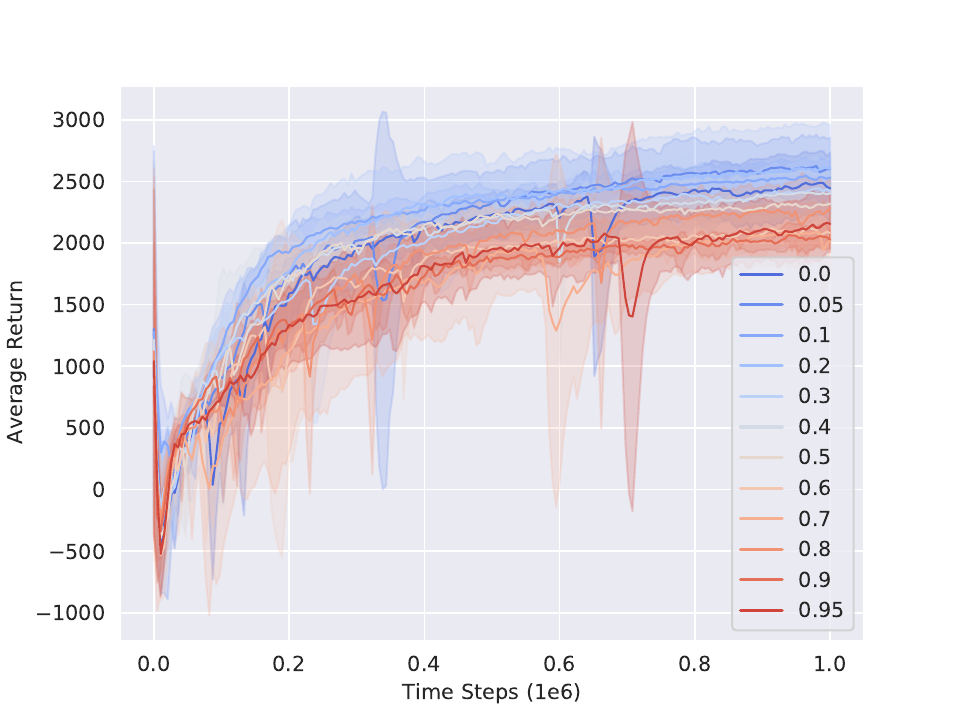}
	}   
	\hspace{-0.3in}
	\subfigure[Hopper]{
		\includegraphics[width=1.8in]{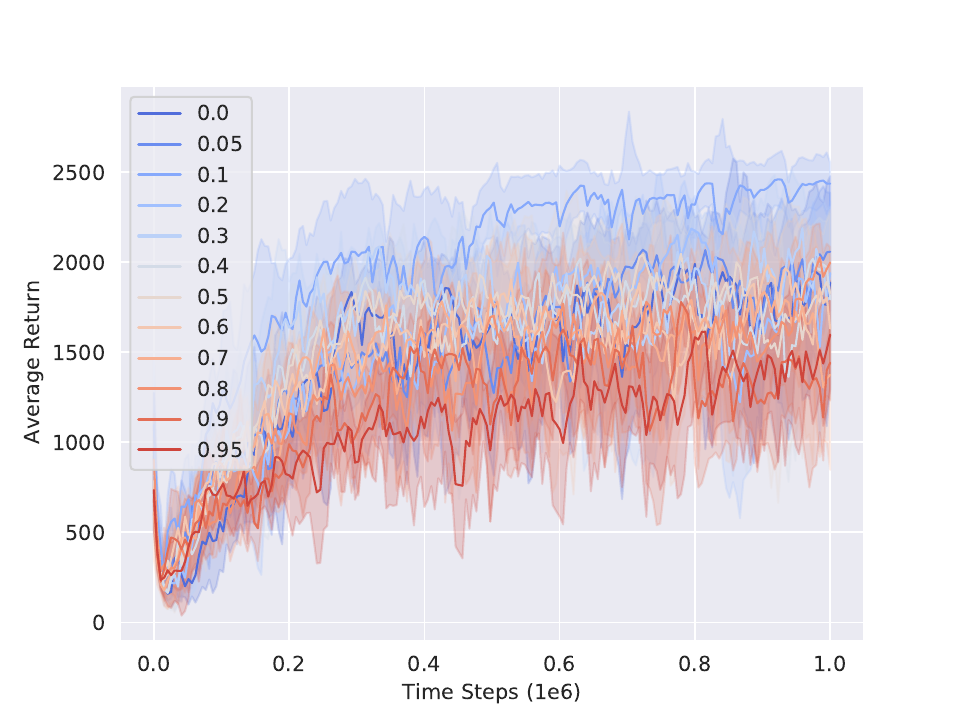}
	}
	\hspace{-0.3in}
	\subfigure[InvertedDoublePendulum]{
		\includegraphics[width=1.8in]{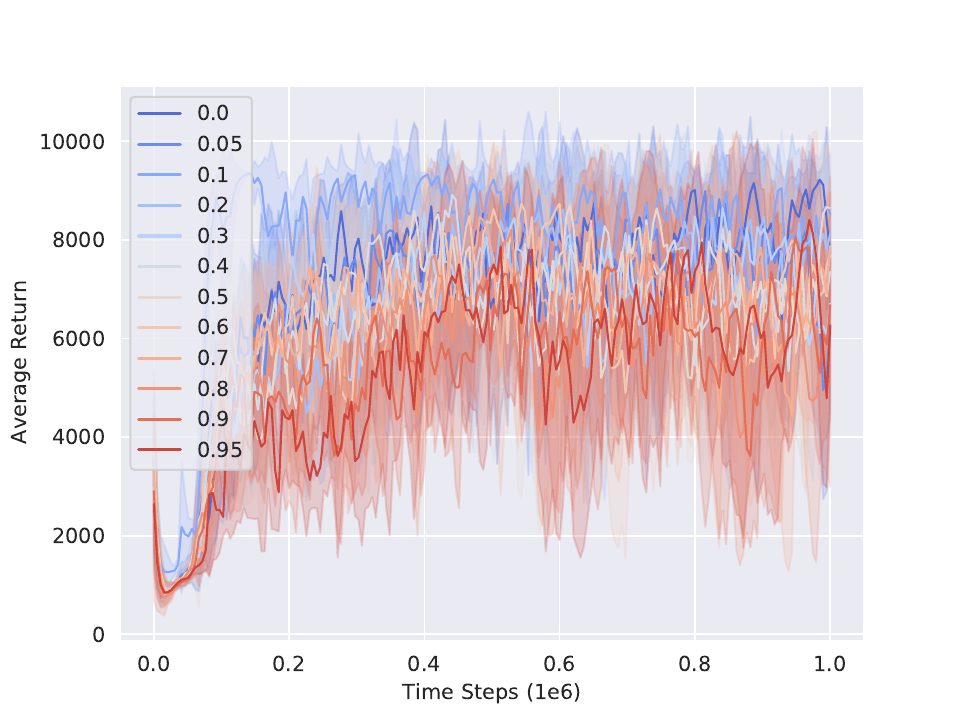}
	}
	\hspace{-0.3in}
	\subfigure[InvertedPendulum]{
		\includegraphics[width=1.8in]{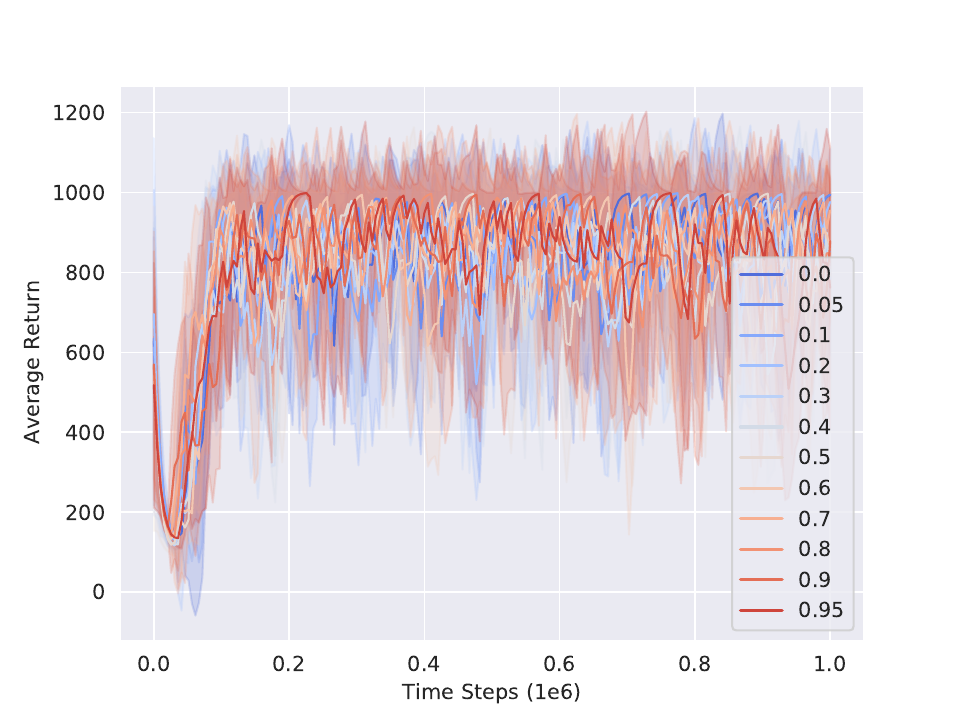}
	}
	\hspace{-0.3in}
	\subfigure[InvertedPendulumSwingup]{
		\includegraphics[width=1.8in]{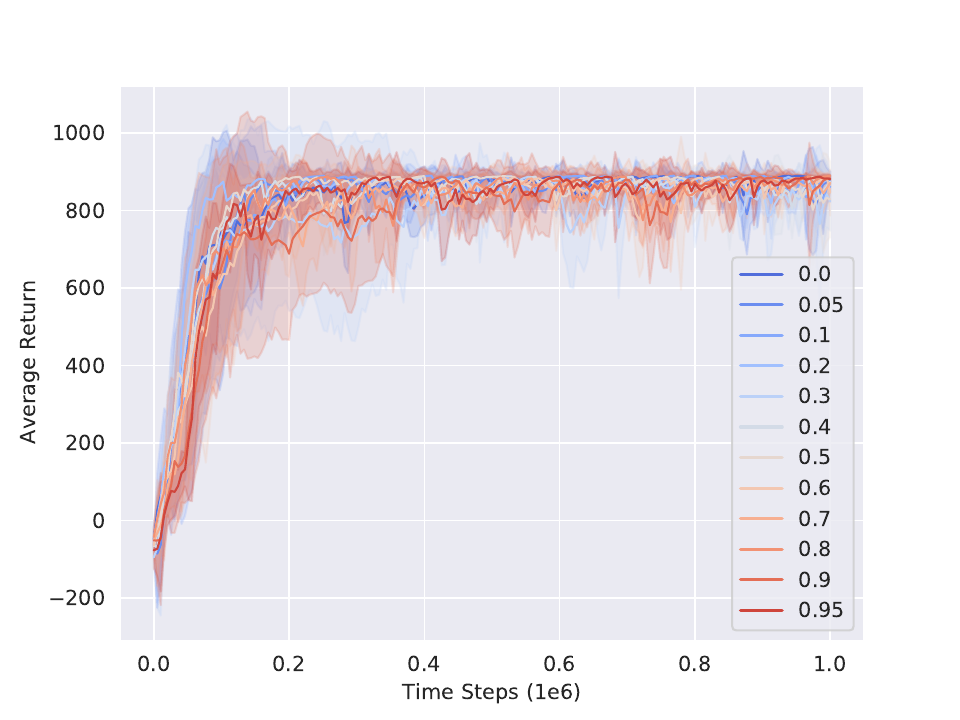}
	}   
	\hspace{-0.3in}
	\subfigure[Reacher]{
		\includegraphics[width=1.8in]{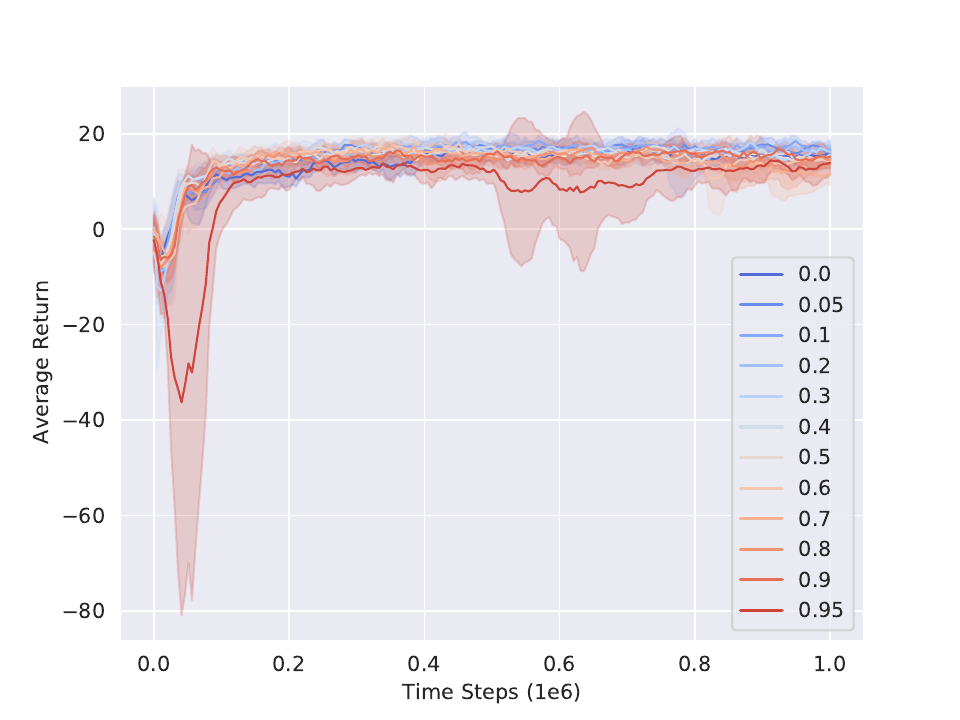}
	}
	\hspace{-0.3in}
	\subfigure[Walker2D]{
		\includegraphics[width=1.8in]{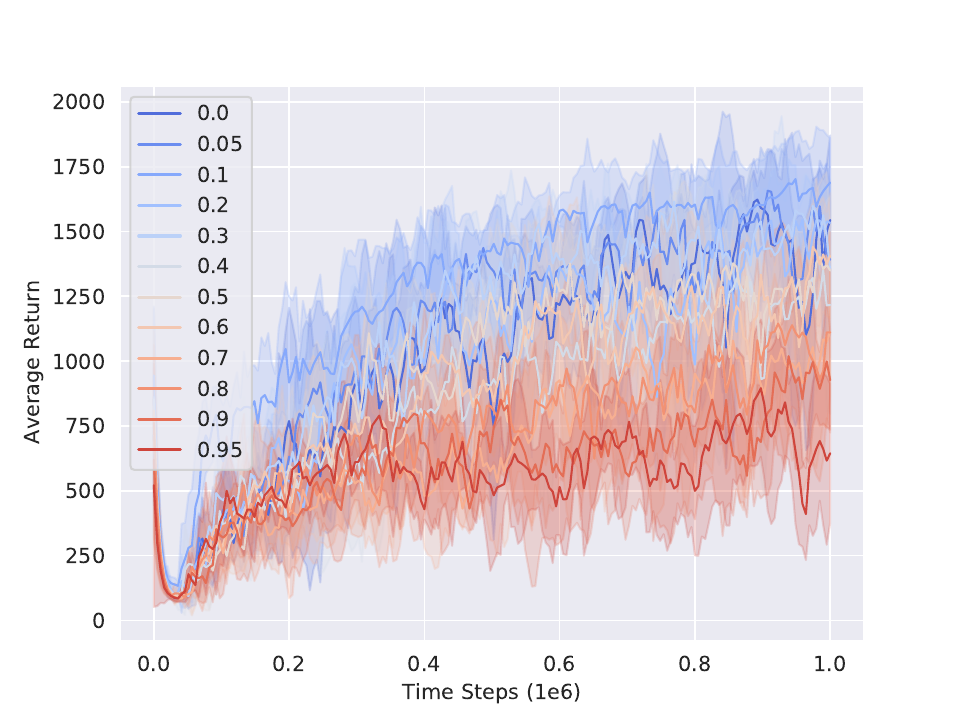}
	}
	\caption{Performance curves for sensitivity analysis on gym PyBullet suite. The shaded area shows a standard derivation.}
	\label{app fig: sensitivity for mepg-ddpg all env}
\end{figure*}

\begin{figure*}[t]
	
	\centering
	\subfigure[Ant]{
		\includegraphics[width=1.8in]{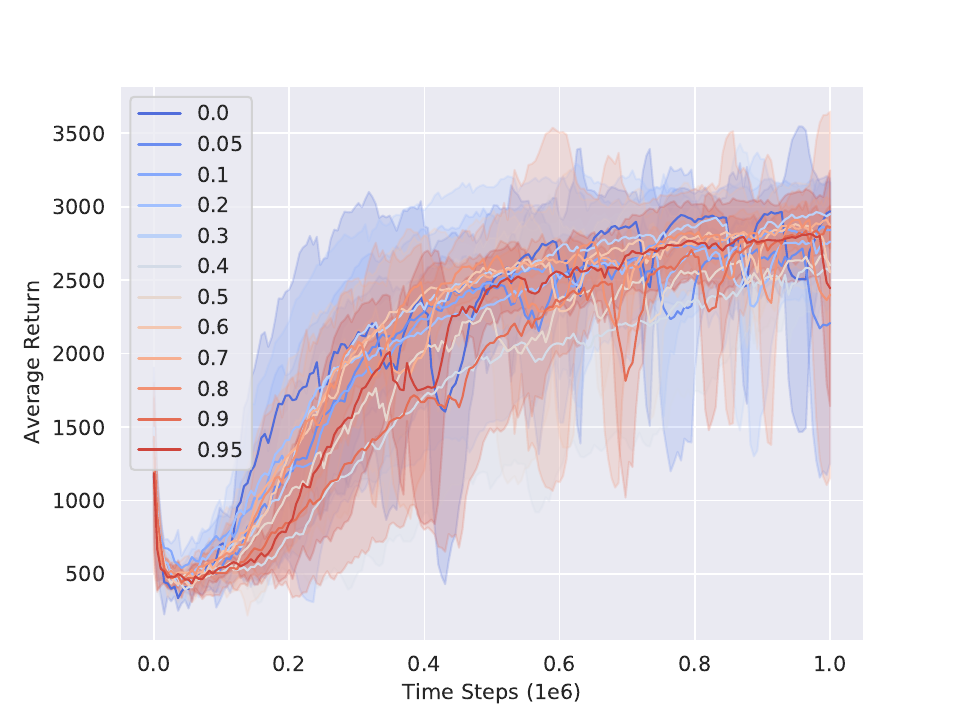}
	}
	\hspace{-0.3in}
	\subfigure[HalfCheetah]{
		\includegraphics[width=1.8in]{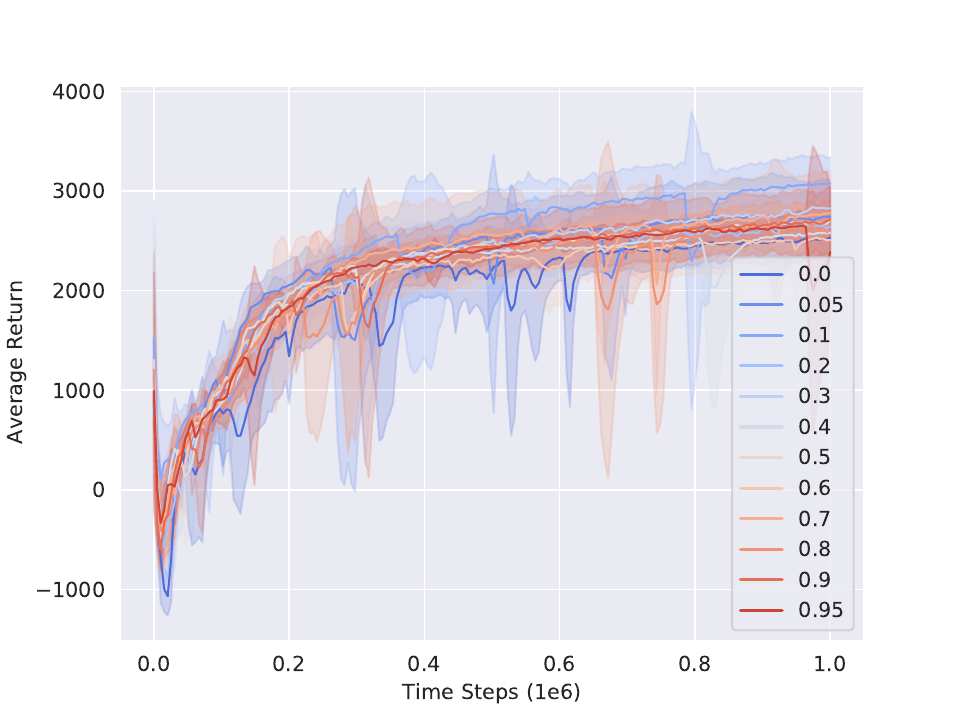}
	}   
	\hspace{-0.3in}
	\subfigure[Hopper]{
		\includegraphics[width=1.8in]{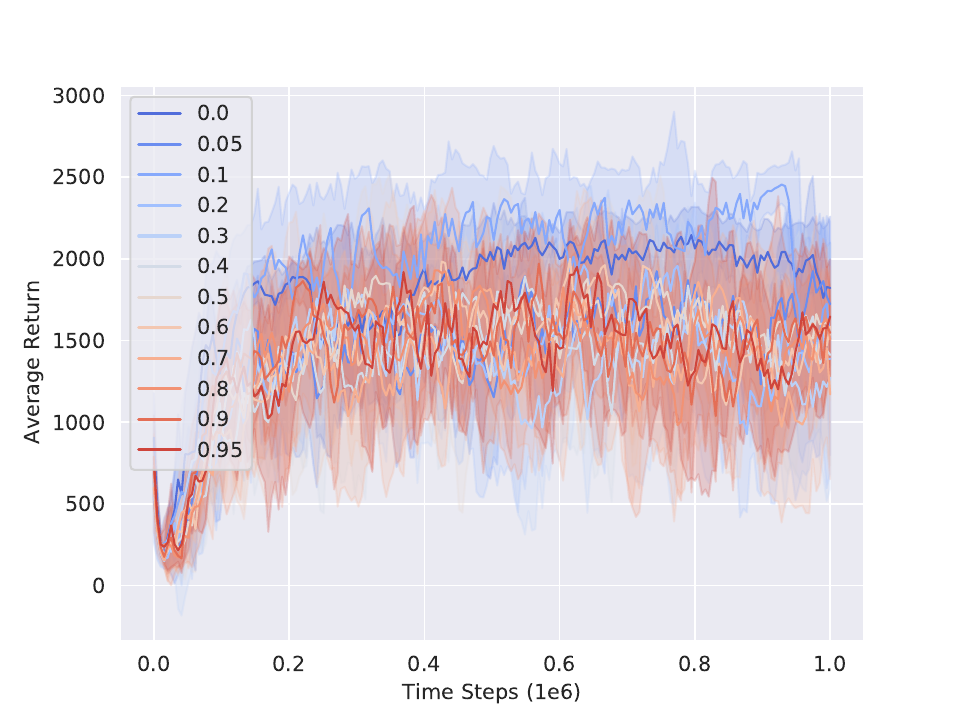}
	}
	\hspace{-0.3in}
	\subfigure[InvertedDoublePendulum]{
		\includegraphics[width=1.8in]{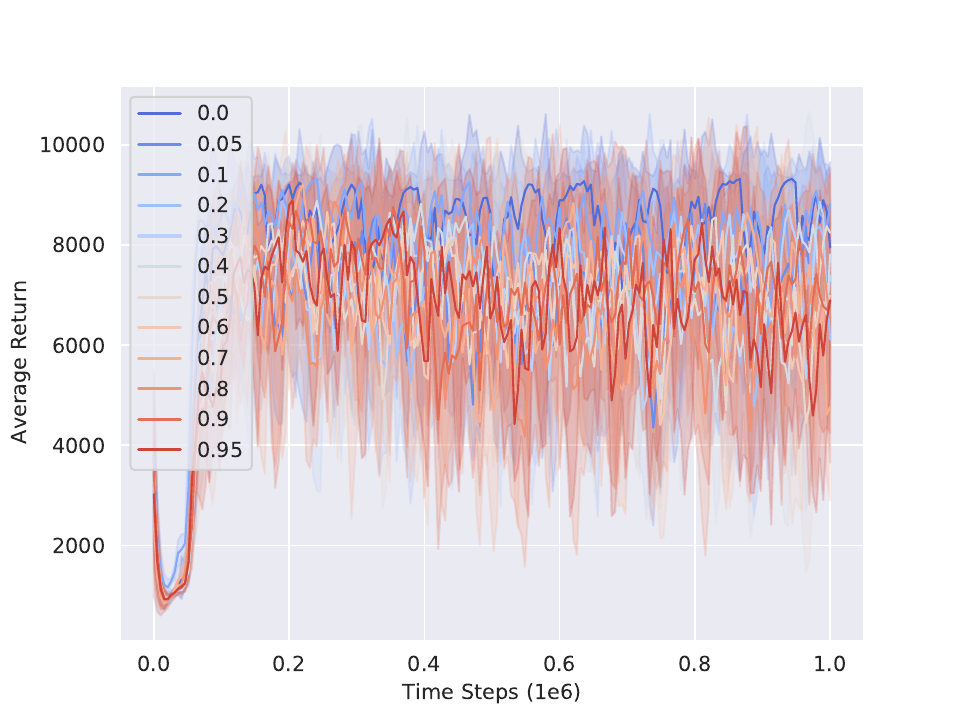}
	}
	\hspace{-0.3in}
	\subfigure[InvertedPendulum]{
		\includegraphics[width=1.8in]{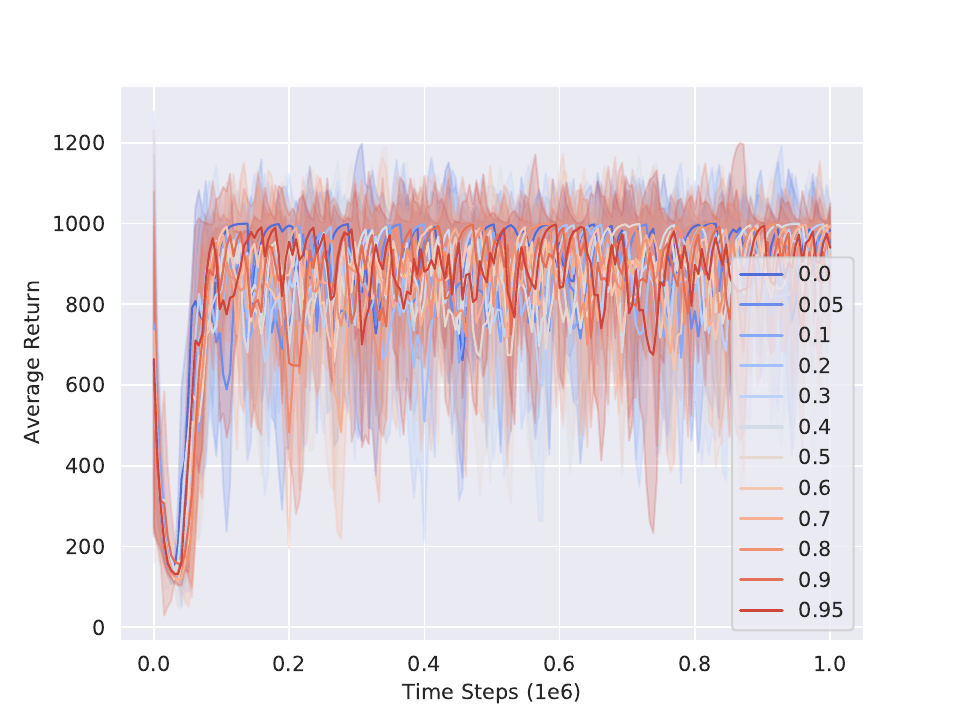}
	}
	\hspace{-0.3in}
	\subfigure[InvertedPendulumSwingup]{
		\includegraphics[width=1.8in]{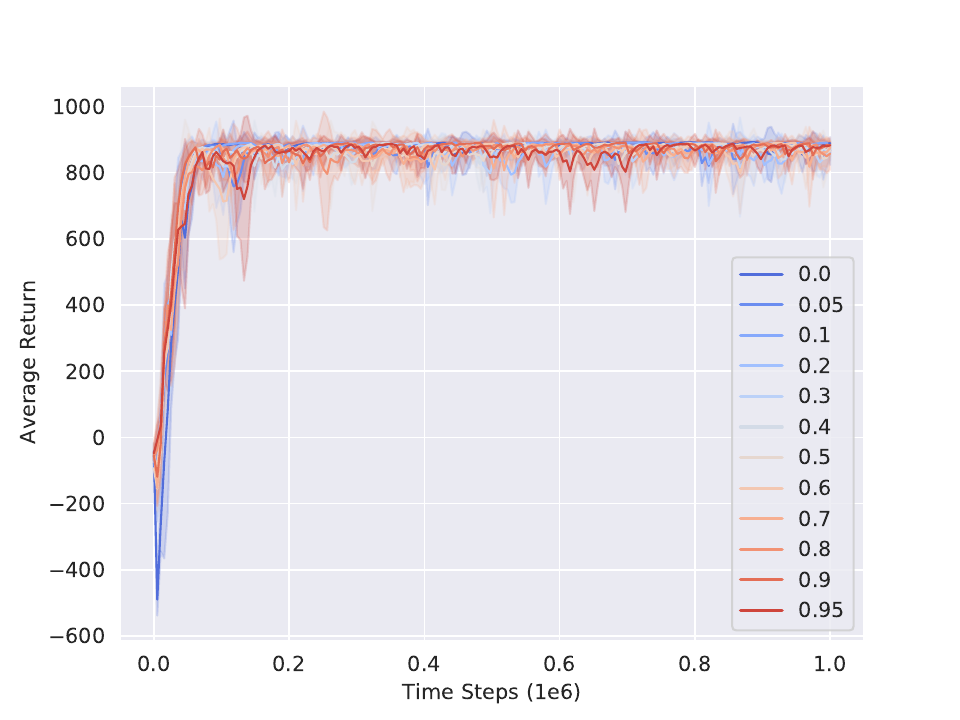}
	}   
	\hspace{-0.3in}
	\subfigure[Reacher]{
		\includegraphics[width=1.8in]{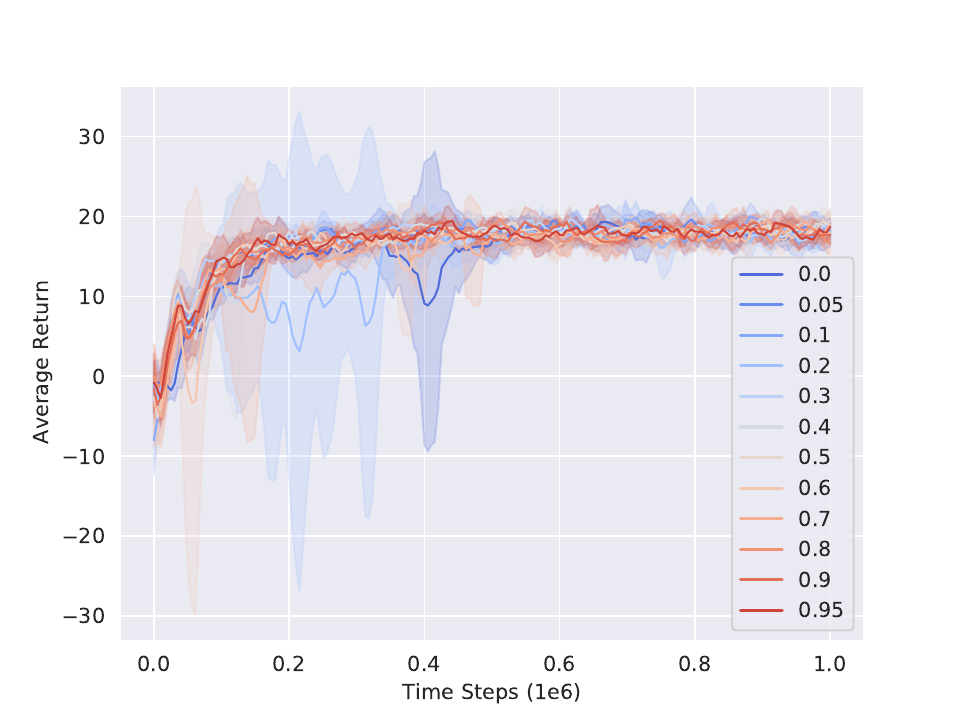}
	}
	\hspace{-0.3in}
	\subfigure[Walker2D]{
		\includegraphics[width=1.8in]{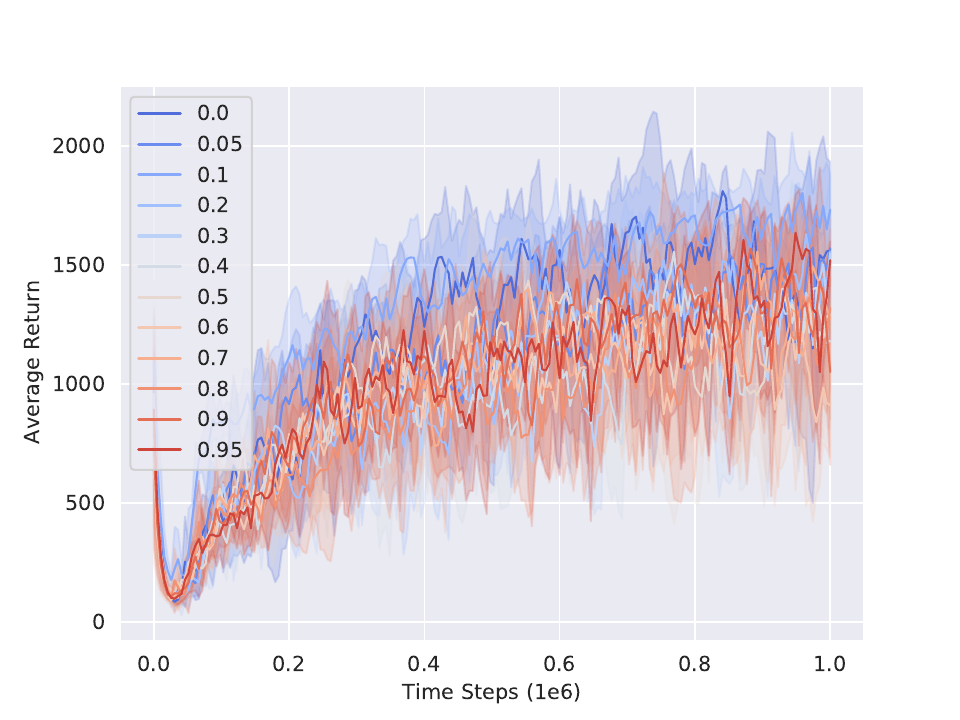}
	}
	\caption{Performance curves for sensitivity analysis on gym PyBullet suite. The shaded area shows a standard derivation.}
	\label{app fig: sensitivity for ME-SAC all env}
\end{figure*}

\textbf{Hyper-paramter sensitivity}. We provide the learning curves for different $p$ in different environments in Figure \ref{app fig: sensitivity for mepg-ddpg all env} and Figure \ref{app fig: sensitivity for ME-SAC all env}. And we also provide the average of top five maximum average returns of one million time steps in Table \ref{app table: sensetivity ME-DDPG all env} and Table \ref{app table: sensetivity ME-SAC all env}.

\subsection{Implementation details and Hyper-parameter setting}
We utilize the authors' implementation of ACE, TD3, SUNRISE ,and REDQ without any modification as we discussed. For the implementation of SAC, we refer to \citep{pytorch_sac}. And we do not change the default hyper-parameters for TD3, SAC, ACE, SUNRISE, and REDQ algorithms. For a fair comparison, we keep the same hyper-parameters to TD3 and SAC implementations respectively. But we only utilize one critic with its target. If the hyper-parameters of ME-SAC and ME-DDPG correspond to the SAC and DDPG, we use the same hyper-parameters. To implement minimalist ensemble consistent Bellman update, we take a $\mathbf{1}$ matrix, then let the $\mathbf{1}$ matrix pass through a dropout layer. The output of the dropout produces a consistent mask. We apply the same mask to the critic and its target.
We give a detailed description of our hyper-parameters in Table \ref{table: hyper-parameters}. 
\begin{table*}[htbp]
	\caption{Hyper-parameters settings for our implementation.}
	\label{table: hyper-parameters}
	\renewcommand\tabcolsep{3.0pt} 
	\centering
	\begin{tabular}{l|c}
		\toprule
		\textbf{Hyper-parameter}&Value\\
		\midrule
		\textit{Shared hyper-parameters} & \\
		discount ($\gamma$) & 0.99 \\
		Replay buffer size & $10^6$ \\
		Optimizer & Adam \citep{kingma2015adam} \\
		Learning rate for actor & $3 \times 10^{-4}$ \\
		Learning rate for critic & $3 \times 10^{-4}$ \\
		Number of hidden layer for all networks & 2 \\
		Number of hidden units per layer & 256 \\
		Activation function & ReLU \\
		Mini-batch size & 256  \\
		Random starting exploration time steps & $2.5 \times 10^4$ \\
		Target smoothing coefficient ($\eta$) & 0.005 \\
		Gradient Clipping & False \\
		Target update interval ($d$) & 2\\
		\midrule
		\textit{TD3} & \\
		Variance of exploration noise & 0.2 \\
		Variance of target policy smoothing & 0.2 \\
		Noise clip range & $[-0.5, 0.5]$ \\
		Delayed policy update frequency & 2 \\
		\midrule
		\textit{ME-DDPG} & \\
		Variance of exploration noise & 0.2 \\
		Variance of target policy smoothing & 0.2 \\
		Noise clip range & $[-0.5, 0.5]$ \\
		Delayed policy update frequency & 2 \\
		Dropout probability ($p$) & 0.1 \\
		\midrule
		\textit{SAC} & \\
		Target Entropy & - dim of $\mathcal{A}$ \\
		Learning rate for $\alpha$ & $1\times 10^{-4}$ \\
		\midrule
		\textit{ME-SAC} & \\
		Target Entropy & - dim of $\mathcal{A}$ \\
		Learning rate for $\alpha$ & $1\times 10^{-4}$ \\
		Dropout probability ($p$) & 0.1 \\
		\bottomrule
	\end{tabular}
\end{table*}

\begin{table*}[h]
	\caption{The average of top five maximum average returns over five trials of one million time steps for various algorithms. The maximum value for each task is bolded. "InvPen", "InvDou" and "InvPenSwingup" are shorthand for "InvertedPendlum", "InvertedDoublePendlum" and "InvertedPendulumSwingup" respectively.}
	\label{app table: performance table 8 results}
	\renewcommand\tabcolsep{3.0pt} 
	\begin{center}
		\begin{tabular}{ccccccccc}
			\toprule
			\textbf{Algorithm}&\textbf{Ant}&\textbf{HalfCheetah}&\textbf{Hopper}&\textbf{Walker2D}&\textbf{InvPen}&\textbf{InvDouPen}&\textbf{InvPenSwingup}&\textbf{Reacher}\\
			\midrule
			ME-SAC &\textbf{2906.98} &\textbf{3113.21} &2532.98 &\textbf{1870.53} &1000.0 &9359.96 &\textbf{893.71} &24.51\\
			ME-DDPG &2841.04 &2582.32 &\textbf{2546.56} &1770.11 &1000.0 &\textbf{9359.98} &893.02 &24.34\\
			SUNRISE &1234.89 &2017.94 &2386.46 &1269.78 &1000.0 &9359.93 &893.2 &\textbf{27.44}\\
			REDQ(Ours) &2238.62 &1757.02 &2238.84 &1092.58 &1000.0 &9359.16 &891.37 &26.02\\
			TD3 &2758.38 &2360.2 &2190.12 &1868.65 &1000.0 &9359.66 &890.97 &24.99\\
			SAC &2009.36 &2567.7 &2317.64 &1776.34 &1000.0 &9358.78 &892.36 &24.13\\
			DDPG &2446.75 &2501.24 &1918.42 &1142.71 &1000.0 &9358.94 &546.65 &24.78\\
			\bottomrule
		\end{tabular}
	\end{center}
\end{table*}


	

\begin{table*}[htb]
	\caption{Run time comparison of training each RL algorithm for eight PyBullet environments. "InvPen", "InvDou" and "InvPenSwingup" are shorthand for "InvertedPendlum", "InvertedDoublePendlum" and "InvertedPendulumSwingup" respectively.}
	\label{app table: run time}
	\renewcommand\tabcolsep{3.0pt} 
	\begin{center}
		\begin{tabular}{ccccccccc}
			\toprule
			\textbf{Algorithm}&\textbf{Ant}&\textbf{HalfCheetah}&\textbf{Hopper}&\textbf{Walker2D}&\textbf{Reacher}&\textbf{InvPen}&\textbf{InvDouPen}&\textbf{InvPenSwingup}\\
			\midrule
			ME-DDPG &47m &44m &42m &45m &37m &36m &36m &36m\\
			TD3 &57m &55m &52m &55m &48m &47m &47m &47m\\
			DDPG &1h 1m &58m &56m &58m &51m &49m &50m &49m\\
			\midrule
			ME-SAC &1h 14m &1h 13m &1h 10m &1h 12m &1h 5m &1h 3m &1h 4m &1h 4m\\
			SAC &1h 17m &1h 15m &1h 12m &1h 15m &1h 8m &1h 6m &1h 6m &1h 6m\\
			REDQ(Ours) &3h 43m &3h 41m &3h 29m &3h 35m &2h 57m &3h 10m &3h 12m &3h 12m\\
			SUNRISE &7h 24m &7h 15m &7h 10m &7h 16m &7h 1m &6h 56m &7h 6m &6h 57m\\
			\bottomrule
		\end{tabular}
	\end{center}
\end{table*}




\begin{table*}[htbp]
	\caption{The average of the top five maximum average returns over five trials of one million time steps for ablation studies. $\pm$ means adding or removing the corresponding component. The maximum value for each task is bolded. "InvPen", "InvDou" and "InvPenSwingup" are shorthand for "InvertedPendlum", "InvertedDoublePendlum" and "InvertedPendulumSwingup" respectively.}
	\label{app table: ablation for me-ddpg}
	\renewcommand\tabcolsep{3.0pt} 
	\begin{center}
\begin{tabular}{ccccccccc}
	\toprule
	\textbf{Algorithm}&\textbf{Ant}&\textbf{HalfCheetah}&\textbf{Hopper}&\textbf{Walker2D}&\textbf{InvPen}&\textbf{InvDouPen}&\textbf{InvPenSwingup}&\textbf{Reacher}\\
	\midrule
	ME-DDPG &2841.04 &2582.32 &2546.56 &1770.11 &1000.0 &9359.98 &893.02 &24.34\\
	ME-DDPG+CDQ &2892.9 &2003.77 &2584.1 &2094.6 &1000.0 &9358.76 &890.76 &15.05\\
	ME-DDPG-DO &2001.56 &2522.97 &2326.2 &1774.61 &1000.0 &9358.73 &892.04 &21.82\\
	ME-DDPG-DU &2610.23 &2575.64 &2330.16 &1784.34 &1000.0 &9359.18 &890.84 &24.08\\
	ME-DDPG-TPS &2623.13 &2605.99 &2328.32 &1675.31 &1000.0 &7522.81 &891.4 &23.93\\
	DDPG &2446.75 &2501.24 &1918.42 &1142.71 &1000.0 &9358.94 &546.65 &24.78\\
	\bottomrule
\end{tabular}
	\end{center}
\end{table*}

\begin{table*}[htbp]
	\caption{The average of the top five maximum average returns over five trials of one million time steps for ablation studies. $\pm$ means adding or removing the corresponding component. The maximum value for each task is bolded. "InvPen", "InvDou" and "InvPenSwingup" are shorthand for "InvertedPendlum", "InvertedDoublePendlum" and "InvertedPendulumSwingup" respectively.}
	\label{app table: ablation for me-sac}
	\renewcommand\tabcolsep{3.0pt} 
	\begin{center}
\begin{tabular}{ccccccccc}
	\toprule
	\textbf{Algorithm}&\textbf{Ant}&\textbf{HalfCheetah}&\textbf{Hopper}&\textbf{Walker2D}&\textbf{InvPen}&\textbf{InvDouPen}&\textbf{InvPenSwingup}&\textbf{Reacher}\\
	\midrule
	ME-SAC &2906.98 &\textbf{3113.21} &\textbf{2532.98} &1870.53 &1000.0 &\textbf{9359.96} &\textbf{893.71} &24.51\\
	ME-SAC+CQD &\textbf{3039.11} &2209.46 &2408.86 &\textbf{2060.8} &1000.0 &9356.73 &890.18 &21.68\\
	ME-SAC-DU &2605.52 &2718.17 &2211.27 &1631.99 &1000.0 &9357.95 &892.2 &24.7\\
	ME-SAC+FIXENT &818.03 &733.59 &1632.85 &843.01 &1000.0 &9358.94 &835.59 &\textbf{26.32}\\
	SAC &2009.36 &2567.7 &2317.64 &1776.34 &1000.0 &9358.78 &892.36 &24.13\\
	\bottomrule
\end{tabular}
	\end{center}
\end{table*}

\begin{table*}[htbp]
	\caption{The number of parameters is given in millions. For all tested environments, we utilize the same network architecture and the same hyper-parameters. The number of parameters differs because different environments have various state and action dimensions, which results in different input and output dimensions of the neural network.}
	\label{app table: number of parameters}
	\centering
	\renewcommand\tabcolsep{3.0pt}
\begin{tabular}{ccccccccc}
	\toprule
	\textbf{Algorithm}&\textbf{Ant}&\textbf{HalfCheetah}&\textbf{Hopper}&\textbf{Walker2D}&\textbf{InvPen}&\textbf{InvDouPen}&\textbf{InvPenSwingup}&\textbf{Reacher}\\
	\midrule
	ME &0.302M &0.297M &0.283M &0.293M &0.271M &0.275M &0.271M &0.276M\\
	ME-SAC &0.226M &0.223M &0.212M &0.220M &0.203M &0.206M &0.203M &0.207M\\
	TD3 &0.453M &0.446M &0.425M &0.440M &0.407M &0.413M &0.407M &0.414M\\
	SAC &0.377M &0.372M &0.354M &0.367M &0.339M &0.344M &0.339M &0.345M\\
	REDQ &1.586M &1.564M &1.489M &1.543M &1.424M &1.446M &1.424M &1.451M\\
	SUNRISE &1.132M &1.117M &1.063M &1.101M &1.017M &1.032M &1.017M &1.036M\\
	ACE &1.614M &1.597M &1.532M &1.577M &1.477M &1.496M &1.477M &1.500M\\
	\bottomrule
\end{tabular}
\end{table*}

\begin{table*}[htb]
	\caption{The average of the top five maximum average returns of ME-DDPG algorithm in a run over five trials of one million time steps for various p-value. The maximum value for each task is bolded. Overall, the difference in performance between the different p-values is not very significant. Relatively speaking, smaller p-values give better performance.}
	\label{app table: sensetivity ME-DDPG all env}
	\renewcommand\tabcolsep{3.0pt} 
	\begin{center}
\begin{tabular}{ccccccccc}
	\toprule
\textbf{p}&\textbf{Ant}&\textbf{HalfCheetah}&\textbf{Hopper}&\textbf{Walker2D}&\textbf{InvPen}&\textbf{InvDouPen}&\textbf{InvPenSwingup}&\textbf{Reacher}\\
\midrule
0.0 &2001.56 &2522.97 &2326.2 &1774.61 &1000.0 &9358.73 &892.04 &21.82\\
0.05 &2641.03 &\textbf{2668.75} &2311.99 &\textbf{1846.5} &1000.0 &9359.11 &891.21 &23.16\\
0.1 &2841.04 &2582.32 &\textbf{2546.56} &1770.11 &1000.0 &\textbf{9359.98} &\textbf{893.02} &\textbf{24.34}\\
0.2 &2616.36 &2645.25 &2326.77 &1747.17 &1000.0 &9359.44 &891.41 &23.14\\
0.3 &\textbf{2791.95} &2481.9 &2443.04 &1807.43 &1000.0 &9359.35 &890.52 &23.36\\
0.4 &2746.82 &2457.72 &2244.52 &1548.22 &1000.0 &9359.14 &890.61 &23.12\\
0.5 &2603.27 &2376.87 &2283.3 &1687.19 &1000.0 &9359.21 &890.7 &23.38\\
0.6 &2512.87 &2174.87 &2169.95 &1631.38 &1000.0 &9359.34 &890.77 &23.01\\
0.7 &2790.01 &2146.34 &2226.71 &1360.51 &1000.0 &9359.31 &890.22 &22.84\\
0.8 &2552.4 &2360.52 &2216.69 &1335.82 &1000.0 &9359.15 &890.97 &23.09\\
0.9 &2427.33 &2119.58 &2029.8 &1266.88 &1000.0 &9359.37 &890.76 &23.1\\
0.95 &2461.24 &2209.9 &1939.54 &1154.81 &1000.0 &9359.24 &890.54 &20.73\\
	\bottomrule
\end{tabular}
	\end{center}
\end{table*}

\begin{table*}[htb]
	\caption{The average of the top five maximum average returns of ME-SAC algorithm in a run over five trials of one million time steps for various p-value. The maximum value for each task is bolded. Overall, the difference in performance between the different p-values is not very significant. Relatively speaking, smaller p-values give better performance.}
	\label{app table: sensetivity ME-SAC all env}
	\renewcommand\tabcolsep{3.0pt} 
	\begin{center}
		\begin{tabular}{ccccccccc}
			\toprule
			\textbf{p}&\textbf{Ant}&\textbf{HalfCheetah}&\textbf{Hopper}&\textbf{Walker2D}&\textbf{InvPen}&\textbf{InvDouPen}&\textbf{InvPenSwingup}&\textbf{Reacher}\\
			\midrule
			0.0 &\textbf{3036.48} &2569.0 &2284.7 &\textbf{1973.73} &1000.0 &\textbf{9359.1} &893.35 &24.32\\
			0.05 &2772.88 &2800.64 &2385.61 &1864.32 &1000.0 &9358.91 &892.12 &24.53\\
			0.1 &2906.98 &\textbf{3113.21} &2532.98 &1870.53 &1000.0 &9359.96 &893.71 &24.51\\
			0.2 &2835.34 &2683.07 &\textbf{2392.05} &1849.91 &1000.0 &9358.78 &892.13 &24.82\\
			0.3 &3018.22 &2875.04 &2308.31 &1796.81 &1000.0 &9358.82 &892.16 &\textbf{25.04}\\
			0.4 &2673.15 &2623.41 &2386.58 &1778.47 &1000.0 &9358.21 &892.18 &25.15\\
			0.5 &2721.36 &2726.1 &2386.39 &1815.28 &1000.0 &9358.95 &892.27 &25.02\\
			0.6 &2919.25 &2554.68 &2332.94 &1744.59 &1000.0 &9358.85 &892.15 &23.79\\
			0.7 &2937.18 &2804.65 &2287.79 &1810.17 &1000.0 &9358.9 &892.18 &24.47\\
			0.8 &2974.49 &2717.76 &2233.29 &1756.7 &1000.0 &9358.5 &892.01 &24.5\\
			0.9 &2927.92 &2737.94 &2358.26 &1811.99 &1000.0 &9358.88 &892.41 &24.54\\
			0.95 &2883.05 &2690.42 &2391.78 &1834.05 &1000.0 &9358.59 &892.38 &24.99\\
			\bottomrule
		\end{tabular}
	\end{center}
\end{table*}

\end{document}